\documentclass{sn-jnl}
\usepackage{apacite} 

\usepackage{placeins}

\usepackage{hyperref}

\usepackage{url}

\usepackage{booktabs}

\usepackage{amsfonts}
\usepackage{amsmath}
\usepackage{amssymb}

\usepackage{nicefrac}

\usepackage{microtype}

\usepackage{fancyhdr}

\usepackage{graphicx}
\usepackage{tabularx}
\usepackage{float}
\usepackage{rotating}
\usepackage{placeins}
\usepackage{comment}

\usepackage{anyfontsize}

\graphicspath{ {.} }

\usepackage{algorithm}
\usepackage{algpseudocode}

\usepackage{parskip}

\newcommand{\captionciteA}[1]{\protect\citeA{#1}}

\begin{document}

\title[Demystifying RL in Production Scheduling via xAI]{Demystifying Reinforcement Learning in Production Scheduling via Explainable AI}

\author*[1]{\fnm{Daniel} \sur{Fischer}
}
\email{daniel.fischer@hsbi.de}

\author[1]{\fnm{Hannah M.} \sur{Hüsener}
}
\email{hannah\_maria.huesener@hsbi.de}

\author[1]{\fnm{Felix} \sur{Grumbach}}
\email{felix.grumbach@hsbi.de}

\author[1]{\fnm{Lukas} \sur{Vollenkemper}}
\email{lukas.vollenkemper@hsbi.de}

\author[2]{\fnm{Arthur} \sur{Müller}}
\email{arthur.mueller@iosb-ina.fraunhofer.de}

\author[1]{\fnm{Pascal} \sur{Reusch}}
\email{pascal.reusch@hsbi.de}

\affil[1]{
\orgdiv{Center for Applied Data Science (CfADS)},
\orgname{Hochschule Bielefeld}
\orgaddress{
\city{Gütersloh}, 
\country{Germany}}}
\affil[2]{
\orgdiv{Department of Machine Intelligence},
\orgname{Fraunhofer IOSB-INA},
\orgaddress{
\city{Lemgo},
\country{Germany}}}


\abstract{
Deep Reinforcement Learning (DRL) is a frequently employed technique to solve scheduling problems. Although DRL agents ace at delivering viable results in short computing times, their reasoning remains opaque. We conduct a case study where we systematically apply two explainable AI (xAI) frameworks, namely SHAP (DeepSHAP) and Captum (Input x Gradient), to describe the reasoning behind scheduling decisions of a specialized DRL agent in a flow production. We find that methods in the xAI literature lack falsifiability and consistent terminology, do not adequately consider domain-knowledge, the target audience or real-world scenarios, and typically provide simple input-output explanations rather than causal interpretations. To resolve this issue, we introduce a hypotheses-based workflow. This approach enables us to inspect whether explanations align with domain knowledge and match the reward hypotheses of the agent. We furthermore tackle the challenge of communicating these insights to third parties by tailoring hypotheses to the target audience, which can serve as interpretations of the agent's behavior after verification. Our proposed workflow emphasizes the repeated verification of explanations and may be applicable to various DRL-based scheduling use cases.
}

\maketitle
\section{Introduction}
\subsection{Background and key concepts}
While the application of Artificial Intelligence (AI) is gaining more attention in many disciplines \shortcite{dwivedi2021artificial,fast2017long, heuillet2021explainability, dwivedi2023explainable}, such as medicine or education \shortcite{chen2023explainable}, the public remains divided on its benefits and risks, and generally lacks a deep understanding of AI and machine learning (ML) in general \shortcite{bao2022whose, gillespie2021trust, liehner2023perceptions}. Concerns exist regarding ethics, transparency and job displacement \shortcite{EUaiact, liehner2023perceptions}, but these do not not stop AI researchers from pushing the boundaries of the field \shortcite{peters2019artificial}.

One area where AI is becoming increasingly relevant is manufacturing and scheduling. Scheduling is defined as a decision-making process that "deals with the allocation of resources to tasks over given time periods" \shortcite[p. 1]{pinedo2012scheduling}. The allocation process can be divided into rules, such as a sequence in which machines process production jobs. Here, one way to tackle scheduling problems is Reinforcement Learning (RL). RL is a branch of ML where an agent interacts with an environment and is trained to behave optimally. \hypertarget{RL}{Typically, RL problems are framed as Markov Decision Processes (MDPs).} An MDP is defined as a tuple $(\mathcal{S}, \mathcal{A}, P, R, \gamma)$. The set of states $\mathcal{S}$ represents all possible situations in which the agent can find itself, where each state $s_t$ encodes information about the environment at a given time step $t$. The set of actions $\mathcal{A}$ includes all possible actions the agent can take in any given state. The transition probability function $P$ defines the probability of moving from one state to another given a specific action, encapsulating the dynamics of the environment. The reward function $R$ assigns a numerical value $r_{t+1}$ to each action taken in a particular state, providing immediate feedback to the agent on the desirability of its actions. Lastly, the discount factor $\gamma$ (ranging between 0 and 1) determines the importance of future rewards, balancing immediate and long-term gains in the agent's decision-making process \shortcite{Sutton2018}. RL agents can function as a decision support tool to find near-optimal job sequences in short computing times, even for highly customized configurations. This capability is a significant advantage for automated planning systems, enabling near real-time (re-)scheduling \shortcite{Grumbach2023Memetic}.

Recently, deep reinforcement learning (DRL) as a subcategory of RL has become a preferred method for tackling complex scheduling problems. \shortcite{Kayhan2021, Grumbach2022Robust}. DRL leverages deep neural networks (DNNs) to approximate the functions that map states to actions and estimate the associated rewards. This combination allows DRL agents to handle high-dimensional state spaces and complex environments, making them particularly effective in scenarios where traditional RL methods may struggle. By integrating deep learning techniques, DRL enhances the scalability and flexibility of RL, enabling it to solve more intricate scheduling tasks \shortcite{Kayhan2021, Grumbach2022Robust}.

Although DRL performs well on a wide variety of tasks, the underlying motives and reasoning of the agent typically remain opaque. This so called black-box nature poses a significant hurdle for companies and users in terms of trust as well as adoption \shortcite{arrieta2020explainable} and therefore requires explanation. 
To gain insight into the black-box of models such as DNN and DRL, the field of explainable AI (xAI) is thus aimed at making these models understandable and more transparent to increase trust and acceptance towards it. At the same time, xAI can shed light on possible bias, robustness of predictions as well logical validation and satisfy other objectives depending on the stakeholder \shortcite{heuillet2021explainability, dwivedi2023explainable, mohseni202124, bekkemoen2023explainable}. Especially in manufacturing and scheduling, xAI is seldom used and existing approaches rarely take into account domain knowledge of target users \shortcite{chen2023explainable, chen2023applications}.

Within the field of xAI, explainable Reinforcement Learning (xRL) \shortcite{puiutta2020explainable} refers to explaining a (D)RL agent’s actions during decision-making. 
xRL is fundamentally tied to interpretable ML and due to this overlap there is less research on xRL in specific. \shortcite{dazeley2023explainable}

For manufacturing and scheduling, xRL on the one hand has to convince the production planners and decision-makers that the agent is trust-worthy and makes reasonable decisions \shortcite{heuillet2021explainability}. On the other hand, companies need to ensure that the model is robust \shortcite{zhang2019robust}. In both ways, it might be helpful to generate explanations that take into consideration the domain, audience (e.g., non-experts in AI), and application of the used AI model, instead of creating predominantly mathematical or data science jargon heavy explanations that are difficult to communicate to various stakeholders \shortcite{heuillet2021explainability, dwivedi2023explainable, bekkemoen2023explainable, chen2023explainable}. 

We schematize this problem by analyzing workflows for xAI in practice. Existing workflows heavily focus the development process and deployment of the explanation model and do not take into account concept drift. Transparent workflows for manufacturing which make intuitive sense to management, engineers, production planners and other stakeholders are rare \shortcite{Clement2023, tchuente2024methodological}.
While approaches of xAI are receiving more scientific interest, a universal solution to making AI – or more narrowly DRL – interpretable has yet to be found \shortcite{heuillet2021explainability}. Additionally, many xAI methods lack falsifiability \cite{leavitt2020towards}.

\subsection{Contribution and Research Questions}
We combine two xAI methods and demonstrate how to create falsifiable explanations tailored to AI laymen in a real-world scheduling scenario. We adapt the holistic approach of \citeA{tchuente2024methodological} to guide empirical investigations with xAI in business applications. Our xAI approach involves formulating hypotheses in natural language based on domain knowledge, analyzing the reward function of the agent, and descriptive statistics of use case data. This is to analyze the plausibility of the xAI results and to ensure that the generated explanations can be effectively communicated within the specific domain.

The scope of the paper is to answer the following questions:
\begin{itemize}
    \item \textbf{(RQ1.1)} How can xAI methods, specifically DeepSHAP and Input x Gradient, be systematically implemented, applied, and validated to describe the decisions of a DRL agent in a real-world flow production context?
    \item \textbf{(RQ1.2)} How suitable are the chosen xAI methods for the specified use case and what are their advantages and disadvantages?
    \item \textbf{(RQ2.1)} How can a workflow be developed to integrate hypotheses derived from domain knowledge with xAI methods for scheduling applications, ensuring falsifiability?
    \item \textbf{(RQ2.2)} How can these xAI-explanations be processed and communicated to stakeholders using domain knowledge?
\end{itemize}
\section{Literature Review}
\label{literature}
We start by laying a common ground on RL in production and scheduling. Afterwards, we introduce xAI methods and workflows before we stress the research gap for falsifiable and systematic xAI approaches in production planning. In the following, we differentiate between frameworks (software tools which \textit{contain} xAI methods) and the methods themselves that can be mathematically formulated. In accompanying literature, proposed xAI workflows are often found under the synonyms "framework", "process", "process model" or "procedure".

\subsection{Reinforcement Learning Applications in Production and Scheduling}
Referring to the definition by \citeA{pinedo2012scheduling}, scheduling as is has to be optimized to meet the expectations of stakeholders.  There are several possible dependent variables, such as lead times, resource utilization, or simply costs.
One way to approach this problem is to solve it heuristically or analytically (e.g., with operations research methods). Although these methods are widely studied and applied, heuristics tend to find only local optima, while mathematical models are only suitable for smaller problems with a simple search space.

RL provides an alternative to heuristic methods, trying to utilize ML to solve problems more efficiently. Because scheduling offers constant feedback for the agent through its sensors, the agent can navigate well in these environments \shortcite{khadivi2023deeprlmanufacturing, sutton2018reinforcement}. Over the years, several novel approaches emerged which are highly suitable for scheduling problems: Q-Learning \shortcite{watkins1992q} and DRL (\hyperlink{RL}{see Introduction}) \shortcite{mnih2015drl} to name a few. The last is especially useful when the state space size is not displayable in a table format as former RL algorithms did.
DRL outperforms classical RL approaches and Q-Learning on a variety of scheduling tasks, see \citeA{khadivi2023deeprlmanufacturing} for an overview. For example, \citeA{SERRANORUIZ2024100582} use OpenAI Gym to model a quasi-realistic job shop scheduling environment using a digital twin based on a MDP and DRL with the proximal policy optimization algorithm, incorporating an observation space with more than a dozen job features and an action space with three heuristic priority rules, which demonstrated superior multi-objective performance compared to traditional heuristic rules.

An advantage of these DNN lies in their ability to model complex structures of the state space. The agent is then able to use this information for its decisions. A major downside is the black-box structure. Even if the reward function is chosen carefully, the resulting actions and the reached solutions seem promising, no conclusions can be drawn about the \textit{reasons} behind the agent's decisions. It is expected that there is a huge information potential hidden in networks used for DRL \shortcite{kuhnle2022explainable}.
In order to reveal these hidden information and enhance the explainability of DRL models, several approaches have been proposed.

\subsection{State of the art xAI frameworks and methods}
Before diving into the xAI approaches, a common terminology has to be laid out as \citeA{palacio2021xai} propose in their framework for unifying xAI.

Following this framework, the \textit{explanation} is the description of information made to create understanding by humans. \textit{How} explanations are created is defined by the xAI \textit{methods}. Here, the methods need to specify what is needed as input and in return what output is produced. 
On the other hand, the \textit{interpretation} is the meaning that is associated with the explanation. This also relates to \textit{why} this meaning is assigned (e.g., a causal link or the context). \shortcite{palacio2021xai}

Other researchers argue that in the field of AI, interpretability ("interpretable AI") refers to a lower complexity of the model so that its components can be analyzed and understood (e.g., small decision trees are interpretable) intrinsically. These types of algorithms are sometimes labelled 'transparent'. While xAI approaches can enhance model transparency already during development, for black-box models that are not intrinsically interpretable or transparent, explainers are oftentimes used. These additional algorithms can explain the outcomes after the original model was executed. This is also referred to as \textit{post-hoc} explainability. 
So, while interpretability may also refer to the complexity, explanations specify the reasoning that lead to a behavior.  Following this definition, an AI-model could be interpretable without being explainable and vice versa. A fully explainable model or system is supposed to create explanations for it's own actions. The overarching concept of this is sometimes termed 'Transparent AI'. \shortcite{mohseni202124,kim2021multi, heuillet2021explainability, milani2024explainable}

In the following, we refer to explanations as the output of the xAI methods and interpretations as the meaning as well as conclusions that can be drawn from them. We will further this later on.

Within xAI approaches, four important differentiations have to be made. There are model specific and model agnostic methods, as well as global and local explanations. Methods can either be applicable to only a specific model and algorithm or be used for all types of ML models. How easily xAI methods can be used for different underlying ML models is sometimes called portability. Explanations can be created for the entire model, only specific predictions (e.g., local or instance explanations for single actions) or both \shortcite{heuillet2021explainability, dwivedi2023explainable, mohseni202124, chen2023explainable, milani2024explainable}.

\subsubsection{The choice of explanation type}
Various xAI methods can produce different explanation types for different use cases or needs (for an overview see \citeA{mohseni202124} or \citeA{chen2023explainable}). The choice what type to use is highly dependent on the target audience, which means that one has to consider the background knowledge of the explainee. An AI expert could be interested in detailed explanations for debugging, while laymen might only want to understand enough to decide whether or not they can rely on the AI system (e.g., via explanations or more transparency on the algorithm) \shortcite{mohseni202124,kim2021multi,heuillet2021explainability,dwivedi2023explainable}. 

Researchers can use xAI to describe how the entire model works, for example by using global explanations \shortcite{mohseni202124, chen2023explainable}. In xRL, there are approaches aimed at helping developers debugging their model by better understanding the states, actions and their outcomes \shortcite{dazeley2023explainable}. Quality control might also play a role. For example, \citeA{klar2024explainable} use policy summarization \shortcite{wells2021explainable}, state value evaluation and perturbation methods to filter out random explanations. This way they ensure that random solutions are not influencing the outcome of the planning of a factory layout. 

Example-based explanations concentrate on explaining selected instances in the data, e.g., by using local explanations for describing a single action of the agent \shortcite{chen2023explainable}. However, single explanations are criticized for being insufficient to explain the workings of complex models \shortcite{leavitt2020towards}.

Other types of xAI include what-if explanations (to show how changes influence the output), contrastive explanations, which are aimed at highlighting why certain outcomes did not happen (e.g., "Why not another action?"), and counterfactual explanations to demonstrate which changes in the input or model would lead to a different outcome (e.g., "What has to be observed for a different action?"). These explanations are considered to be easily understandable for humans, but they can be challenging to use in large state spaces or in some cases need specifically generated features. \shortcite{dazeley2023explainable, mohseni202124, kim2021multi, chen2023explainable}

Explaining the reasons for a specific outcome based on an input is termed a \textit{why} explanation \shortcite{mohseni202124}. Within why-explanations, feature-based techniques show how the input features change the model output by assigning importance values \shortcite{dwivedi2023explainable}. 

In the next sections, we will explain some xAI methods in detail. This is not intended to be a complete overview, which is not the scope of this paper.

\subsubsection{xAI Methods}
One of the ways to categorize xAI methods for deep learning is into intrinsic, perturbation- and gradient- or backpropagation-based \shortcite{chen2023explainable, kamath2021explainable}.
We first introduce decision trees, a popular and intrinsically interpretable algorithm that is broadly used. Then, we outline SHAP as well as several perturbation- and gradient-based xAI methods.

\paragraph{Decision Trees}
Decision trees are widely used and offer explainability out of the box \shortcite{kotsiantis2013decision}.  
Starting with the work of \citeA{breiman1984cart} the foundation of modern decision trees has been laid. With his CART (classification and regression tree) the rules discovered by the algorithm are displayable in a transparent form. Decision trees are highly versatile, e.g. for regression and classification. In the original CART algorithm, on each split the information gain has to be maximized which relies on information entropy \shortcite{quinlan1986induction} and the concept of information content \shortcite{shannon1948mathematical}.

Decision trees have been used in a wide range of domains, including stock market \shortcite{wu2006effective}, marketing \shortcite{kim2001application}, image classification \shortcite{yang2003application} and scheduling \shortcite{Portoleau2024}.

\paragraph{SHAP (SHapley Additive exPlanation)}
\label{SHAP}
SHAP is a framework introduced by \citeA{lundberg2017unified}. 
The authors show that multiple xAI methods such as Lime \shortcite{ribeiro2016should} and DeepLIFT \shortcite{shrikumar2016not} are additive feature attribution methods. 
Because of the complex nature of blackbox models, (local) explanation models \textit{g} approximate the original model \textit{f} to make a prediction \textit{f(x)} based on an input \textit{x} interpretable. Oftentimes inputs are simplified \( x' \) and can be mapped back into the original input space using a mapping function x = hx(\( x' \)). In the class of additive feature attribution methods, the explanation model can be written as a linear function: Starting from $\phi_0$, the model output where all simplified inputs are missing, the simplified input features (binary variables) are being attributed an effect $\phi_i$ if they are present. Then, all feature attributions are summed, giving an approximation of the original model prediction. \shortcite{lundberg2017unified}

When the values $\phi_i$ used are Shapley values from cooperative game theory \shortcite{shapley1953value} three properties (local accuracy, missingness, and consistency) for meaningful explanations are satisfied \shortcite{lundberg2017unified}. Note that Shapley values are permutation-based \shortcite{shapley1953value}.
SHAP values are "Shapley values of a conditional expectation function of the original model" \shortcite[p. 4]{lundberg2017unified}, where every feature is assigned an importance value that shows how the expected model prediction changes when conditioning on that specific feature. The starting point of this additive attribution is the so called base value E[f(z)], which represents the expected prediction if all feature values were unknown. \shortcite{lundberg2017unified}

\[g(z^{\prime})=\phi_{0}+\sum_{i=1}^{M}\phi_{i}z_{i}^{\prime},\]

where $z' \in \{0,1\}^M$, $M$ is the number of simplified input features,  and $\phi_i \in \mathbb{R}$.

When input features are correlated or when dealing with non-linearity, the order of adding features to the expectation is relevant; however, to simplify computation independence between features is usually assumed in SHAP \shortcite{lundberg2017unified}. Due to permutation and especially if features are dependent, this can result in individual SHAP values for unrealistic data points \shortcite{aas2021explaining, kamath2021explainable}, which imposes a limitation of the SHAP methods.

While the exact computation of SHAP values for real-world data is difficult, the authors propose model-agnostic and model-type-speciﬁc methods for approximation, whereas using model-speciﬁc information can improve computational performance. Using SHAP, both local and global explanations can be created \shortcite{dwivedi2023explainable}. SHAP has previously been used for xAI in manufacturing \shortcite{chen2023explainable}. 

\paragraph{DeepLIFT}
DeepLIFT \shortcite{shrikumar2016not} is an additive feature attribution method \shortcite{lundberg2017unified}. It recursively explains a DNN's predictions by comparing the original activation of each neuron to a reference (background) value for every input and assigning an importance score to individual parts of the input \shortcite{lundberg2017unified, shrikumar2016not}. In DeepLIFT, the contribution of one neuron to another is defined to satisfy the properties of summation to delta and linear composition, which is achieved using the backpropagation rules proposed by the authors (see \cite{shrikumar2016not} for more). A limitation of this method lies in the need to determine an appropriate reference.

In the prediction of weaning from mechanical ventilation, DeepLIFT has been used to ensure that a convolutional NN makes predictions using clinically important features \shortcite{jia2021prediction}.

\paragraph{DeepSHAP}
The DeepSHAP explainer is speciﬁc for deep learning models, utilizing their compositional structure \shortcite{lundberg2017unified}. It builds on DeepLIFT \shortcite{shrikumar2016not}. 
If the input features are independent, the neural network is linearized, shapley values are chosen as attribution values and the reference value is taken to be E[x], DeepLIFT approximates SHAP values. Making use of DeepLIFT’s form of back-propagation (multipliers \textit{m}), DeepSHAP combines the SHAP values for a feature i and the prediction y for the network components (e.g., \( f_3 \)) – that can be solved analytically if linear – into SHAP values for the entire network. \shortcite{lundberg2017unified}

$$\phi_i(f_3, y) \approx m_{y_{if_3}}(y_i - E[y_i])
$$

A limitation of DeepSHAP lies in its dependence on the background input that must be chosen to compute the mean prediction \shortcite{fernando2019study}. Stability improves with larger background sample sizes \shortcite{yuan2022empirical} .

Recently, DeepSHAP has been used to explain the results of a DNN in the field of condition monitoring for hydraulic systems \shortcite{keleko2023health}.

\paragraph{Input X Gradient}
The method Input X Gradient (or \textit{gradient*input}) 
works by taking the partial derivatives of the output with respect to the input and multiplying them with the input itself \shortcite{kindermans2016investigating}:

$$R_{i}^{c}(x) = x_i \cdot \frac{\partial S_c(x)}{\partial x_i}$$

where $R_{i}^{c}(x)$ is the contribution of each given input $x_i$, $S_c(x)$ is the output function and  $\frac{\partial S_c(x)}{\partial x_i}$ is the partial derivative of the output function with respect to the input. While the partial derivative itself can provide information regarding how an infinitesimal change in the input influences the output, Input X Gradient goes one step further \shortcite{adebayo2018sanity}. Also being able to detect gradient saturation is a huge benefit in contrast to the plain gradient consideration.

One upside of Input x Gradient lies in its monitorability. For attribution values, decision boundaries and critical thresholds can be defined. For example, if the attribution for a certain action contradicts the 68–95–99.7 rule \shortcite{wooditch2021normal}, an alert may be sent out.
Input X Gradient is sensitive to scaling \shortcite{sundararajan2016gradients}. \citeA{leavitt2020towards} mention that Input x Gradient can lead to unreliable explanations when considering single neuron activations. False claims could be made by making the false assumption that a single neuron may be \textit{causally} responsible for a specific action of the agent. This disregards the complex interactions between the neurons in the whole network. 

\citeA{chatterjee2024exploration} use Input X Gradient to colour specific areas of the lung by patients eventually suffering from COVID-19. \citeA{ozer2023explainable} also use this technique for x-ray images. 

\paragraph{Layer-wise relevance propagation}
Layer-wise relevance propagation (LRP) uses heatmaps to show single pixel contributions to an output \shortcite{bach2015pixel, binder2016layer}. It works by a special layer-to-layer backpropagation method which shows the contribution of every neuron to the prediction \shortcite{MONTAVON20181}. Each layer receives so called relevance scores from the succeeding layer and redistributes them proportionally to its inputs. There is not a method of choice in calculating these scores, so the modeller is free in determining what "relevance" means in the given context. The $\alpha \beta$ rule, among others, tries to balance relevance between positive and negative contributions \shortcite{kohlbrennerlrp}.

\citeA{yang2018explaining} use LRP in a medical domain to generate several therapy suggestions. They validate the explanations with experts to support their findings. \citeA{arras2016explainingpredictionsnonlinearclassifiers} apply LRP to text classification and identify important words in documents. 

Since we are explaining an RL-model, we researched general xAI methods as well as specific methods in xRL. Different xRL methods exist, but as \citeA{heuillet2021explainability}  point out many of those are not applicable to RL in real-world scenarios: One reason being that in RL, there are specific assumptions, algorithms and environments with different constrains to consider. The authors propose to focus research on global xAI approaches. 
In fact, we find approaches in literature where DRL results were explained using general xAI methods (such as SHAP or even decision trees) that are not specific to RL \shortcite{10566218}.
Nonetheless, the next section reviews state of the art xRL methods.

\subsubsection{xRL}
\citeA{xiong2024xrl} state that in xRL there a multiple different approaches and respective methods: One can explain the model logic, the reward function (by decomposition or shaping), the states and the tasks. Specifically for explaining the states, post-hoc methods such as SHAP may be used.

\citeA{milani2024explainable} propose to categorize xRL into methods for feature importance (e.g., local state–action relationships), training and MDP (f.e., what objective is prioritized), as well as policy-explanations (global long-term behavior explanations), so that they align closer with RL-logic.
For feature-importance, there are different explanation formats and various approaches. Methods can range from using surrogate models to make the policy interpretable, to using inherently interpretable models that can structure policy information (such as decision trees), up to simply explaining what states were important for an action using visualization (e.g., saliency maps) or natural language.
Regarding training and MDP, the authors split xRL methods into three categories. Some methods use the transitions and explain the agent's behavior with causal models. Another method is reward decomposition, where – before training – the reward function is constructed in a way that comprises different reward types with a meaning (e.g., to explain that a specific reward component is more important than another for certain actions). However, the authors point out that this technique only works for specific Q-learning algorithms. Another method in xRL regarding the training and MDP consists of identifying and inspecting data points that were most influential in the training of the agent. 
Lastly, policy-explanation methods either inspect long-term agent behavior for important states in training or for clustered similar states. Another approach is converting the NN that represents the policies into an understandable representation.

The meta review by \citeA{bekkemoen2023explainable} sheds light on xRL explanations and stakeholder needs. They develop a new taxonomy which differentiates between three types of xRL agents: Interpretable Agents, Intrinsic Explainability and Post Hoc Explainability. Interpretable agents are intepretable out of the box with a single function approximator. Intrinsic explainability describes preparation of the RL system before training to make it explainable. Post hoc explainability aims at describing the agent \textit{after} training. Because intrinsic and post hoc explainability overlap, they developed additional criteria for differentiating these agents. After categorizing agents, they map several question categories (by "How (to)?", "What?", "Why (not)?", "What if?) onto the discovered agents and their capabilities. They also stress the importance of the validity period and method used.

Regarding intrinsic explainable DRL, in the field of network slicing, \citeA{10283684} developed a method that uses the reward hypothesis to create an xAI reward for the agent by combining SHAP and entropy values \shortcite{shannon1948mathematical} to increase decision certainty. This xAI reward is paired with the task-related reward in the training phase for a resource allocation optimization problem. The authors show superior performance of the explainable DRL agent in comparison to standard DRL.

Asides from explainability, \citeA{xiong2024xrl} highlight that there is a lack of standardized evaluation metrics to compare xRL methods. They categorize evaluations into subjective (e.g., user's understanding and trust) and objective. In their xRL-Bench, they focus on the latter, specifically on fidelity (e.g., faithfulness: Does the explanation match agent's logic?) and stability (consistency of explanations). This is assessed by either masking or perturbing important and unimportant states and inspecting deviations in the model outputs before and after. 
\citeA{milani2024explainable}  also discuss metrics for xRL methods and review which metrics are used in literature. Interestingly, they find that fidelity and understandability of xRL are rarely assessed, while visualizations are most popular. Though not quantitative, visualizations are often utilized to illustrate interpretability. 

\subsubsection{Frameworks for xAI workflow and design decisions}
Asides from the methods themselves, we must also consider the procedure of generating xAI in a complex business setting. This includes the workflow, design choices as well as other aspects to look out for. We draw on frameworks from both the xAI community as well as other domains.

\citeA{leavitt2020towards} point out that approaches of xAI, especially for deep learning, often rely on visualization or single examples and lack falsification via hypotheses, quantifiability and human as well as general verification of validity. \citeA{chen2023explainable} also highlight that xAI results need to be validated. This is something that should be kept in mind in the workflow and design of explanations, in order to ensure a scientific standard.

\citeA{mohseni202124} propose a framework for xAI design and evaluation. In their nested model, the outer layer focuses on the outcomes. When designing an xAI system, general considerations have to be made first: What is the goal, the target group and what is supposed to be explained? This also involves determining how to evaluate the xAI system to measure if the expectations were met.
In the next layer, the process of explaining has to be decided on with the user in mind. This includes aspects like the explanation formats and the amount of details to include, while also evaluating the usefulness of the chosen explanations.
The inner layer is the type of xAI method itself. When black-box models are explained with ad-hoc algorithms, fidelity to the original model plays a role. The trustworthiness of the underlying model should also be evaluated.

In their framework for design and analysis of xAI, \citeA{kim2021multi} outline the historical development of explanations and point out that explanations should to be scientific or at least causal. 
\citeA{dazeley2023explainable} also highlight the importance of causality and introduce the Causal xRL Framework to create causal explanations in RL. They draw on two existing theories.
First, \citeA{dazeley2021levels} suggest that xRL can take place at different orders and that higher level explanations should be incorporated for acceptance of AI systems. Zero-order explanations only consider how an input resulted in an output (base case in xAI). First-order explanations that consider an agent’s objective to maximise a reward signal are also relevant. Explaining the agent's intentionality (e.g., the objectives behind the behavior) might improve understanding of the agent, in comparison to only giving zero-order explanations \shortcite{dazeley2021levels, dazeley2023explainable}.
Secondly, due to the temporal sequence of transitions in RL, \citeA{dazeley2023explainable} propose to provide causal explanations. They draw on the Casual Explanation Network by \citeA{bohm2015people} , which aims at explaining intentionality, cause and reason of behavior as well as implications for the future.
In our setting, the Causal xRL Framework can be understood as following: The agent is expected to make production decisions based on two objectives (e.g., goals). For this, the agent observes the state (perception, zero-order) and depending on the attributes of the features one objective might be prioritized (disposition, first-order). To fulfill the objective (first-order), the agent will then make an action that will in return cause an outcome (zero-order).
If an agent has more than one objective the explanation of disposition plays a role, which \citeA{dazeley2023explainable} point out as an outlook for future research. 
Goal-driven xAI approaches are currently an emerging research field \shortcite{dazeley2023explainable}. However, it is mainly a focus in autonomous agents and robots, while research on xRL and reactive agents in particular is limited to policy retention without including domain knowledge \shortcite{sado2023explainable}.

\citeA{shi2023explainable} widen the circle of people involved by incorporating domain experts and researchers together. Their approach is suited for manufacturing and consists of a prototype of a task guiding system. With a graphical user interface they summarize information about different models and agent behaviour. This makes it easier for third parties to interact with the explanations. \citeA{langer2021what} build upon that and create a model to focus on the different \textit{stakeholder desiderata}. Their model classifies different stakeholders and their demands regarding xAI explanations. Different stakeholder classes like deployers, regulators, users and "affected" are presented. One key insight are the highly versatile expectations of the different stakeholders. These expectations are already affected by the different domain knowledge of the parties. The intersection of these interests has to be found to provide a reasonable xAI model.

Asides from the xAI literature, we also researched workflows from other domains.
From a broad perspective, using a unified, \textit{general} framework which is well-founded in the literature makes sense. The \textbf{CR}oss \textbf{I}ndustry \textbf{S}tandard \textbf{P}rocess for \textbf{D}ata \textbf{M}ining framework (CRISP-DM) by \citeA{wirth2000crisp} provides a holistic view on all internal departments and decision processes involved for deploying a data mining model. 
However, it does not focus on xAI models specifically and does not contain two layers for the explained model and the explaining model. The model is put to production in the Deployment stage. After that, the framework stops. CRISP-DM is thus not feasible to guide xAI projects after deployment and to verify their explanations. It is also not suitable for environments where a lot of the constraints change and the model has to be adapted. 
Several researchers tried to improve on that.

\citeA{tchuente2024methodological} specifically reviewed xAI in business settings and proposed a workflow for xAI in business applications, which is discussed in detail in our approach \ref{systematic workflow}. As limitations of their workflow, the authors highlight that it has to be adapted to specific contexts and further work is need regarding robustness and validation of explanations. Later on, we directly take up these limitations with our approach.

\subsection{Research Gap}
We did not find many established frameworks for xAI in scheduling. The amount of xRL methods suitable in real-world scenarios is limited \shortcite{bekkemoen2023explainable, heuillet2021explainability}. Also, many xAI methods we reviewed were suitable for debugging, but not for non-AI experts. Here, the questions arises which xAI methods may be applied to describe the decisions of a DRL agent in a real-world flow production context (RQ 1.1)?
How can they systematically be implemented, applied, and validated \shortcite{chi2022quantitative}? 
Every use case poses different challenges for the developer to consider. Explanations may vary according to the questions which arise. Identifying the most suitable xAI method as well as weighing its (dis-)advantages is crucial (RQ 1.2).

Many efforts in xRL only provide zero-order explanations, which are not comprehensive enough to create trust and acceptance in the AI system and only few approaches include utilizing an agent's objectives or dispositions to create casual explanations \shortcite{dazeley2021levels, dazeley2023explainable}. Adding domain knowledge to xRL is important \shortcite{milani2024explainable}, but considering it in xAI approaches for manufacturing is rare \citeA{chen2023explainable}. Here, the questions arise how domain knowledge as well as context can be included into an explanation. How can xAI-results be processed and presented to stakeholders using this knowledge (RQ 2.2)?

In the xAI community, researchers have criticized the lack of unified terminology \shortcite{palacio2021xai} as well as falsification and missing validity checks \shortcite{leavitt2020towards, chen2023explainable}. How can falsifiablity be ensured (RQ 1.1, 2.1)?

We fill this gap by developing a workflow based on hypotheses (RQ 2.1), utilizing the domain knowledge (RQ 2.2) of the users and knowledge of the agent's workings. The framework builds upon existing xAI methods and -frameworks and focuses on real world scheduling applications (RQ 1.1, 1.2).
\section{Approach}
In the following section, we illustrate our real-world use case and the scheduling problem formulation, as well as the data set were are working with. Then, we state our xAI workflow with the aim to fill the research gap. 

\subsection{Preliminaries}
To gain a deeper understanding of our use case, we now formulate the scheduling problem and the MDP to address it. 

\subsubsection{Real-world case and scheduling problem formulation}
\begin{figure}[ht!]
    \centering
    \includegraphics[width=0.77\textwidth]{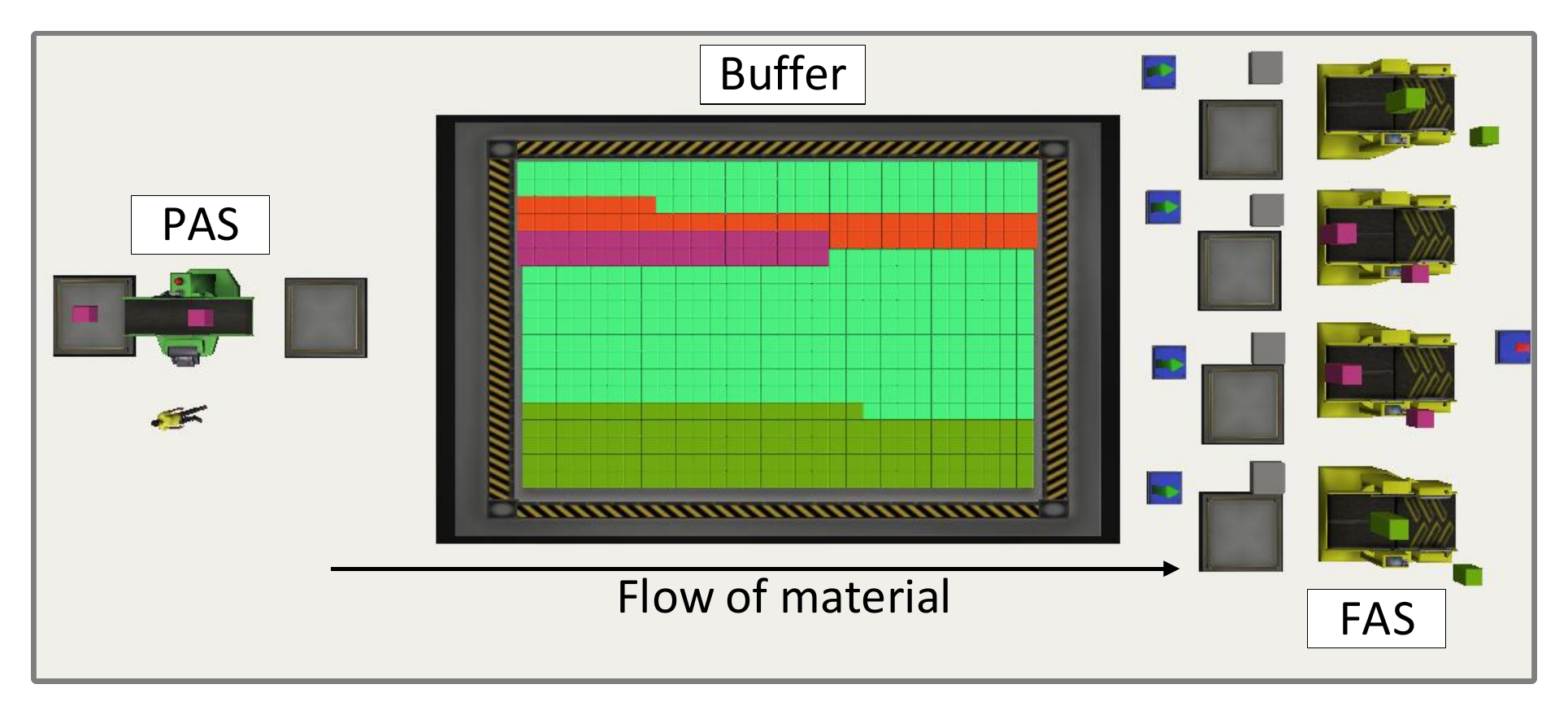}
    \caption{Model of the considered two-stage flow production system by \captionciteA{müller2024reinforcement}.}
    \label{fig:sim_model}
\end{figure}
\FloatBarrier

The systematic application of xAI techniques is examined in the context of a real-world use case at a large German manufacturer of household appliances. In previous publications, an extensive scheduling model, along with a specialized DRL agent, has been investigated, implemented and tuned \shortcite{müller2024reinforcement,müller2024smaller}. The considered manufacturing process involves a two-stage flow production system. As visualized in Figure \ref{fig:sim_model}, the shopfloor consists of a pre-assembly stage (PAS) and a final assembly stage (FAS). In the PAS, a single station produces eight types of semi-finished products (hereinafter referred to as products), which are then finished at one of four possible FAS stations. Between these two stages, there is a limited intermediate buffer where the products are temporarily stored. According to \citeA{müller2024smaller}, the key model components of the extended permutation flow shop can be defined semi-formally as outlined below. For a comprehensive mathematical description, refer to \citeA{müller2024reinforcement}.

\begin{itemize}
    \item Given a set of specific products to be finalized in the FAS, each based on a standardized product of eight different types produced in the PAS.
    \item All PAS and FAS stations can process only one product at a time and all processing times are deterministic.
    \item Each product must be finalized at exactly one of four FAS stations. In this context, each FAS has a predefined schedule containing the sequence of single products to be manufactured.
    \item A FAS station can only start finishing the product when it is available in the central buffer. If the next required product is not available in the central buffer, the FAS station will incur idle times until it becomes available.
    \item In the PAS, all products required by the FAS that are not initially available in the central buffer must be produced. 
    \item Switching from one product type to another may cause sequence-dependent setup efforts in the PAS.
\end{itemize}

Given these constraints, two conflicting objectives are optimized in lexicographical order: first, minimizing idle times in the FAS, and second, minimizing setup efforts in the PAS. This problem is addressed by determining a central decision variable: the sequence in which products are loaded into the production system in the PAS. This decision is managed by the existing agent, which will be described in detail in the following section.

\subsubsection{Existing RL Approach}
The real world setting stated above needs to be formalized. Several approaches are feasible here. For example, the environment could be modelled using operations research methods. Because of RL-agent's abilities to efficiently find local optima, we chose DRL. The scheduling problem is modelled as an MDP as presented below. For a detailed explanation, refer to \cite{müller2024reinforcement}.

\paragraph{State Space}
The state space is encoded as a vector and comprises following elements:
\begin{itemize}
    \item \textbf{next\_24h\_demand\_prod} for all products: Specifies the demand of the related product for the next 24 hours, taking into account the amount left in the buffer. 
    \item  \textbf{end\_of\_planning\_period\_demand} for all products: Specifies the demand of the related product for the planning period left, taking into account the amount left in the buffer.
    \item \textbf{buffer\_content\_duration\_prod} for all products: Specifies how long the amount of an product type in the buffer will suffice to meet the demands of the FAS if this type is no longer produced in the PAS.
    \item  \textbf{buffer\_fill\_level}: Specifies the fill level of the buffer, i.e., the amount of all products in relation to its total capacity.
    \item  \textbf{last\_prod\_type\_is} for all products: Specifies the last type of products manufactured, represented in one-hot encoding. 
\end{itemize}

\paragraph{Action Space}
We define the action space to be discrete. It consists of 8 actions representing the 8 different proudct types with 0 centered start (e.g., action 7 = agent recommends to build product 8). Each time the agent selects an action, a lot of 50 units of the corresponding product type is produced. 

\paragraph{Reward Function}
The primary objective of the agent is to avoid idle times. Rather than penalizing idle times directly, we penalize \textit{criticality}, which we define to be the ratio of $\texttt{next\_24h\_demand\_prod}$ to $\texttt{buffer\_content\_duration\_prod}$. This has proven to be significantly more effective for training, as criticality provides a richer learning signal compared to idle times \cite{müller2024reinforcement}. Furthermore, we penalize the agent if $\texttt{buffer\_content\_duration\_prod}$ for any product type falls under a threshold of 30 minutes, which encourages the agent to maintain a certain margin for each required product type.
The other objective is to minimize setup efforts. Therefore, we add setup efforts weighted by a factor directly to the reward function.

\paragraph{Domain Randomization}
To make the agent decisions more robust, domain randomization (DR) has been used \shortcite{müller2023bridging}.
DR consists of the idea to present not one but many environment distributions to the agent at each episode. The agent is forced to adapt to a wider spectrum of scenarios, making its policy more robust.
In the concrete use case, six different weeks (environments) were presented to the agent at random. Each week had distinct characteristics regarding demand and buffer sizes.

\paragraph{Dataset}
The agent made scheduling decisions based on representative synthetic data that were carefully designed to reflect realistic scenarios. These data were obfuscated to preserve the integrity of the study and protect the interests of the project partner, ensuring no sensitive or proprietary information was compromised. After giving the input values, which consist of checkpoints for the weeks, the agent and environment was initialized and while the agent acted in the environment, we saved the state and action pairs. The generated data frame consists of 103 rows and 35 columns, corresponding to the actions of the agent and the state space.

We started our investigations with an already trained NN; therefore, we consider the whole dataset as a test set. Thus, we achieve the same attribution values for each run through the data in Input X Gradient. Settings where the network is trained again on new data may lead to the necessity to use cross validation and to fit confidence intervals for the attribution values to achieve robustness.
We want to note that we had to rebuild the NN, because the original was not available to use in the xAI methods due to it being a custom class.

Now, how can we systematically explain the data of the agent from our use case using xAI?

\subsection[Domain-knowledge hypotheses]{Domain-knowledge hypotheses to combine falsification, interpretation and communication of xAI}
 As we have pointed out, many efforts in xAI, and xRL specifically, provide zero-order explanations and only few include utilizing an agent's objectives or dispositions to create casual explanations \shortcite{dazeley2021levels, dazeley2023explainable, kim2021multi}. Also, domain knowledge is not considered adequately in the context of manufacturing (RQ 2.2) \shortcite{chen2023explainable}. Thus, we want to utilize the reward function and pair it with domain knowledge \shortcite{mohseni202124,kim2021multi,heuillet2021explainability,dwivedi2023explainable, milani2024explainable} to create 'higher-order' explanations \shortcite{dazeley2023explainable}.  Note that following \citeA{palacio2021xai} the (causal) context and the 'why' behind an AI system's behavior would be part of the interpretation and not the explanation itself. Also, we want to take into account the importance of falsification in xAI \shortcite{leavitt2020towards}. 

We postulate that all of these aspects can be combined in a cohesive way. In order to make explanations interpretable and to ensure user's trust, we need to add context as well as knowledge of the agent's workings \shortcite{dazeley2023explainable, mohseni202124}, while keeping the target audience in mind \shortcite{mohseni202124,kim2021multi,heuillet2021explainability,dwivedi2023explainable,chen2023explainable}. Importantly, these are all information we can put together before the explanation is created: If we have a (causal) understanding of what a trained agent should do in a specific use case and situation based on domain knowledge and the agent's objectives (decomposing the reward into meaningful statements), we can formulate a hypothesis \shortcite{ mohseni202124, dazeley2023explainable,leavitt2020towards, milani2024explainable}. For example, we could say: \textit{In the production line, it is important to keep products in buffer. In production scheduling, an agent is punished if the buffer content is too low. Therefore, in a situation where buffer level is low, the agent should produce more products to fill the buffer}. This hypothesis can then be applied to an explanation created with an xAI method (e.g., a method showing that a specific feature was important for the behavior of the agent). If the hypothesis matches the explanation, we conclude that the AI system operates how we expected and we have already created a humanly comprehensible interpretation that involves context and causality (e.g., the relevant feature indicated that buffer content was low, which is why it was relevant in the decision of the agent to fill the buffer). 
If the hypothesis does not match the explanation created by the xAI method, we have detected an error and need to revise the AI system or integrate further context knowledge. By doing this, we combine falsification using hypotheses \shortcite{leavitt2020towards} with (causal) interpretation \shortcite{palacio2021xai, kim2021multi}  and adding higher-levels for trust and acceptance \shortcite{dazeley2023explainable}.  At the same time, we include the background knowledge of the users who are domain experts \shortcite{mohseni202124,kim2021multi,heuillet2021explainability,dwivedi2023explainable, milani2024explainable}. This specific combination is a novel approach for xRL in a real-world business setting to the best of our knowledge. Thereby, we pursue to develop an approach for xAI in business and logistics, specifically using production scheduling as an example, where experts want to quickly understand whether the output of the AI system aligns with their domain knowledge or not.

\subsection{The choice of xAI methods for our use case}
Regarding the choice of xAI methods, one has to consider the goal and the target audience \shortcite{mohseni202124,kim2021multi,heuillet2021explainability,dwivedi2023explainable}.
In our use case, we are neither interested in interpreting static components of the model, nor are we trying to create a self-explaining model as is a general goal in some xAI research \shortcite{mohseni202124,kim2021multi}. We are using an existing DRL-model to explain (post-hoc) why the agent made it's decisions.

It is important to highlight that debugging or explaining the technical details of the entire model is not suitable for our target audience \shortcite{mohseni202124,dazeley2023explainable}. Our approach is aimed at domain experts in production scheduling with user trust and acceptance in mind. The goal is to provide a simple explanation to understand why the DRL agent created a specific production plan. This is why we do not use contrastive or counterfactual explanations, but instead focus on why the specific production decisions were made – leaving what \citeA{mohseni202124} term 'why'-explanations best suited.

In our use case, features – such as the demand or buffer content – are key components in scheduling; therefore, it seems intuitive to use them to create domain-based explanations. Which is why we chose feature-based xAI techniques to show how the input features influence the model output by assigning importance values \shortcite{dwivedi2023explainable}. 

For RL in real-world scenarios, \citeA{heuillet2021explainability} propose to use broad xAI approaches. We follow this suggestion by utilizing model-agnostic xAI frameworks, namely SHAP and Captum.

To sum up, 'why'-explanations concentrating on the features in the data and model-agnostic xAI frameworks are most suitable for our use case.

In order to provide alternative explanation-styles, which can be helpful for the explainee \shortcite{kim2021multi},  we chose one method – DeepSHAP – that primarily relies on visualization, and another method – Input X Gradient – that is more quantifiable and can be presented in a table-format. We create explanations for every class of action, instead at looking at single instances.

For the interpretation of SHAP values, we follow the available SHAP documentation and \citeA{DBLP:journals/corr/abs-2104-10505}. Attribution values returned by Input X Gradient can be interpreted as standard attribution values. Positive attribution values increase the chance of a specific action being chosen, while negative attributions decrease the probability \shortcite{das2020opportunities}.

\subsection{The choice of a systematic workflow}
\label{systematic workflow}
Our approach uses the workflow by \citeA{tchuente2024methodological}, who provide a robust structure for guiding empirical investigations with xAI in business applications in general (holistic approach). 

The workflow by \citeA{tchuente2024methodological} is specifically adapted to xAI models and anticipates changes in the assumptions of the base model. They include a wide range of business parties in their process to mitigate reservations of sceptics.

\begin{figure}[ht!]
    \centering
    \includegraphics[width=1\linewidth]{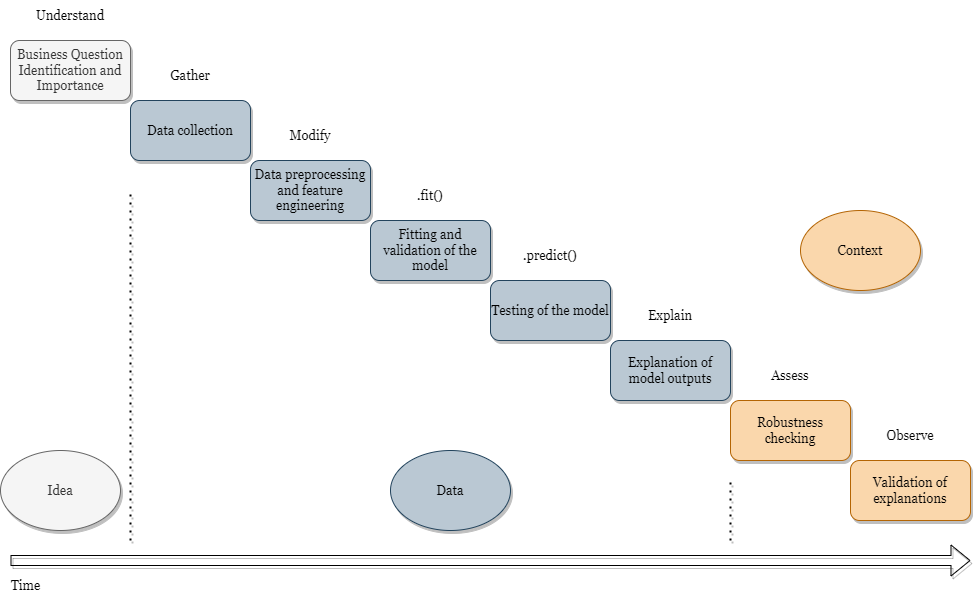}
    \caption{Framework for xAI in business by \captionciteA{tchuente2024methodological}. Idea, data and context are presented as three clusters.}
    \label{fig:tchuente}
\end{figure}
\FloatBarrier

In the following we describe the framework more thoroughly as a basis for our approach to the research questions.

\subsubsection{Business question identification and importance scoring}
The framework by \citeA{tchuente2024methodological} starts with the identification of the business question and its importance (i.e. to the stakeholders). Important (target-)variables may be identified. In order to achieve this, several techniques may be used. It is important to verify that the results stems from a collaboration of many stakeholders to ensure a wide range of opinions and expertise is included in the questions to be answered. The process may be led by engineers or data scientists to ensure that the questions are in fact answerable by algorithms at hand and that the desired objectives are realistic. Lastly, the objectives can be enriched with the sources of the data to help speeding up the next phase. A proper example in this context could be the business question, how the lead time of an exemplary product can be reduced and how these results can be communicated to the stakeholders. Here, the target variable is the lead time. Independent variables could be production times on several machines, dummy variables indicating special product configurations etc. It is important to evaluate what persons may be involved and to include their expertise. 

\subsubsection{Data collection}
After clarifying the desired objective and variables, the data collection starts. Incorporating Data Engineering here might help with the collection. It makes sense to store batch data at disk and to create a pipeline for data streams to guarantee that it is available straight away when needed. The raw data might originate from databases, data warehouses or even pdf-documents. We refer to the widely known ETL process for further reading \shortcite{vassiliadis2002conceptual}.

\subsubsection{Data preprocessing and feature engineering}
Before we can make the data available to the model, we have to handle missing values \shortcite{tchuente2024methodological}. The pipeline used here must be transparent and documented to ensure that third parties can follow which data related decisions had been made \shortcite{chakraborty2017interpretability}. These can highly influence the results of the model and explainer, i.e. if important structures in the data are deleted.

\subsubsection{Fitting and validation of the model}
More important than training the model and validating it, the process of choosing the right model and its communication is key \shortcite{nyawa2023transparent}. \citeA{tchuente2024methodological} emphasize the process of evaluating which model is the most precise. This highly technical decisions may not be interesting to stakeholders at first sight, but may offer problem misconceptions or valuable insights for others. Communication is key even in this phase. At the end of this thought process, the adequate model or model ensemble is found. Explaining its limitations and interpretability is also crucial. 

Another possibility is to explain a model which is already live. In this context, one may directly assess which explainers may be suitable. 

\subsubsection{Testing the model}
This step includes testing the model on the test set. When splitting the dataset into training-, validation- and test set, stakeholders should ensure that the data parts are still representative. 

\subsubsection{Explanation of model outputs}
In this step, the modeller is choosing the explanation method to be used. If only a whitebox model has been chosen in the latter step, this step can be skipped. 
There are several methods available for explaining the model results derived (see chapter xAI methods). The approach chosen has to reflect the objectives the stakeholders want to achieve and the data formats which are at hand. Several explainers only support tabular data or images. Orchestrating several explainers to provide more robust explanations may be beneficial. \citeA{doshi_velez2017towards} point out that the opinion of domain experts and practitioners might be useful here, because they can verify at first glance if results should be carried in the next phase. The formatting of the results should be comprehensible for every party involved and may be embellished for other applications.

\subsubsection{Robustness checking}
If the explanations hold at first glance, one must check them more thoroughly. Robustness in this context means that the explanations provide consistent results. Unfortunately, inconsistent results are possible \shortcite{slack2019fooling} depending on the model explained. Consistency can be verified by double-checking the explanations of different approaches to see if they reach the same conclusions \shortcite{senoner2021using}. Additional strategies relying on measuring the ability of the explainer to emulate the behavior of the original model (fidelity, \shortcite{chi2022quantitative}) or using classic metrics like accuracy are possible. 

\subsubsection{Validation of explanations}
If the results are robust, they have to be validated by domain experts. Validity can include the degree of applicability in practice. In the end result, the results should "make sense" to every party involved in the whole process. 
Because the assumptions made at the deployment of the model forfeit eventually (concept drift), it is important to iterate over the last two steps in a regular interval.
\section{Results and Implications}
We begin by examining the use case to develop hypotheses from a domain-knowledge perspective. These hypotheses are then tested using selected xAI methods. Additionally, we discuss the advantages and disadvantages of the chosen methods. Based on our findings, we propose an adapted workflow and outline the limitations of our process, offering directions for future research. The workflow stems from the approach by \citeA{tchuente2024methodological} and is inspired by their phases Idea, Data and Context.

\subsection{Initial exploration of the use case}
To develop a basis for comprehensible and expertise-based hypotheses, an analysis must be carried out in the specific domain. For this, we analyze the data using an exploratory data analysis \shortcite{tukey1977exploratory}. The chosen action of the agent can be considered the dependent variable in our dataset, while we treat the other variables as features.

\subsubsection{Most manufactured products}
\begin{figure}[H]
    \centering
    \includegraphics[width=0.75\linewidth]{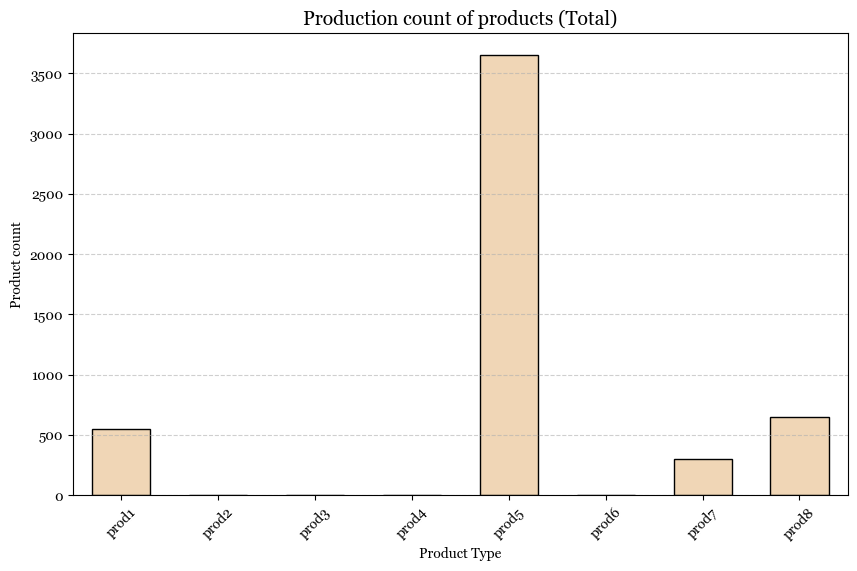}
    \caption{Bar plot of the total amount of products that were produced. Product five has been produced the most. Some products have not been produced at all in the given dataset.}
    \label{fig:Produced_aggs}
\end{figure}

Product 5 was produced the most, followed by product 8, then product 1. Product 7 was produced the least (Fig \ref{fig:Produced_aggs}). The most produced product 5 appeared more than five times as often as the second most produced product 8. The dataset is thus imbalanced regarding product 5 (the dependent variable).

A potential thesis to draw here is that the variables concerning the products that were produced should play a more important role in the xAI methods than the ones concerning those products that were not produced. Variables relating to products that were not produced may still be part of the reasoning for the agent to make a production decision. However, the reason why certain products were produced should be reflected in the importance of the variables concerning these products, due to the domain-specific objectives (e.g., products are produced when there is demand). If an product was produced but none of the variables relating to its demand and buffer seem to play a role, there might be an underlying logical issue.

\subsubsection{Last product type}
\begin{figure}[H]
    \centering
    \includegraphics[width=1\linewidth]{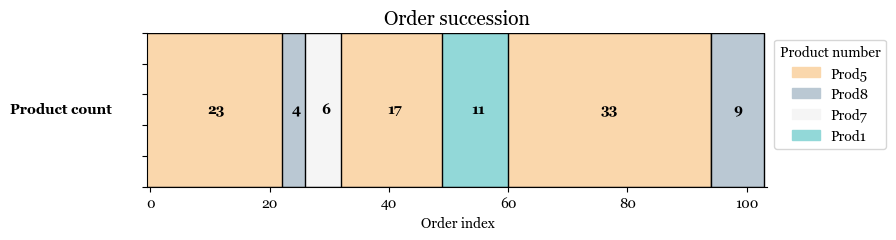}
    \caption{Order of produced product lots. The x-axis shows the order of the products. Product 5's production has been interrupted three times. 33 lots of product 5 have been produced uninterruptedly. This is the longest unbroken sequence.}
    \label{fig:order}
\end{figure}

The last product type feature is a dummy encoded variable (0 or 1) indicating if the predecessor is the same product as produced before. For instance, if \texttt{last\_prod\_type\_is\_prod5} equals 1, the product produced one step before was product 5. It thus encodes the ordering of the products (Fig. \ref{fig:order}). Product 5 has been produced first followed by short periods of products 8 and 7. Product 1 also breaks the production cycle of product 5 at index 50. Product 8 is being produced again a second time at the end.

Besides avoiding idle times in the FAS, the second objective of the agent is to minimize setup efforts in the PAS. Both have to be balanced out, even though idle times are slightly prioritized. As a consequence, we should see that the agent does not constantly change between production of products (unless that is necessary due to an empty buffer or wrong buffer content), which would lead to sequence depended setup efforts. 
We see that this is the case for product 1 and product 7; however, the agent produces product 5 and product 8 with breaks in between. Here, the other main objective of the agent – to avoid high criticality and keep a safety margin – should be of relevance. In certain cases, the agent might need to switch between production of products (at the cost of increasing setup efforts), if criticality of another product is too high.

The insight derived from this is that a) because of the objective of minimizing setup efforts \texttt{last\_prod\_type\_is\_prodX} = 1 increases the probability that the same product will be produced again. However, b) if the criticality of another product is high, the agent switches to producing this product and \texttt{last\_prod\_type\_is\_prodX} = 1 of the current product \texttt{X} will lower the probability of it being produced again (see Table \ref{tab:criticality_analysis}). The objective of avoiding criticality is then of bigger concern than minimizing setup efforts.
If both are relevant at the same time, the effects can be contradictory. For example, minimizing setup efforts will lead to production, even when there is low criticality. High criticality in contrast will interfere with minimizing setup efforts.

\subsubsection{Buffer fill level}
\begin{figure}[H]
    \centering
    \includegraphics[width=0.75\linewidth]{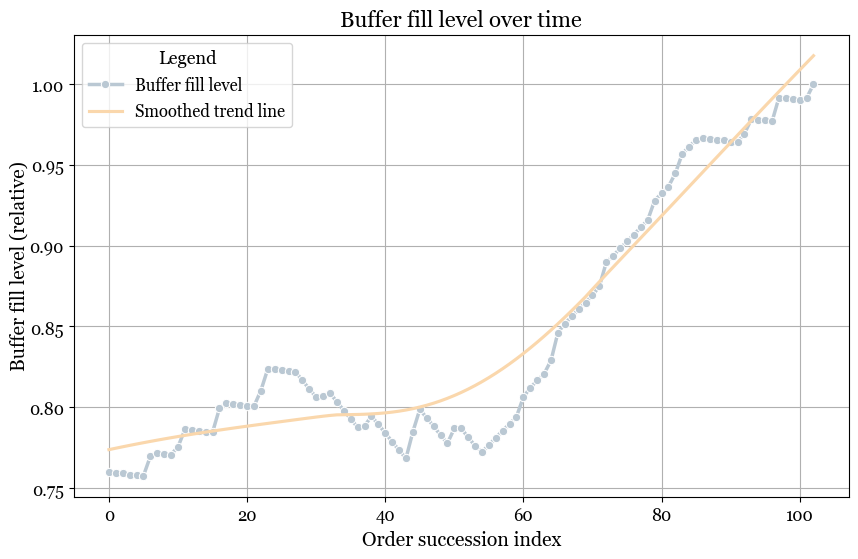}
    \caption{The buffer fill level over 113 decisions of the agent. After index 50 there is a clear upward trend. The trend line has been fitted with a locally weighted liner regression model (\texttt{frac = 0.66}). Note that the y-axis starts at 0.75. Reason: The throughput of the FAS decreases from index 60, so that the PAS fills the buffer with future relevant products. The critical phase in which utilization and setup efforts are balanced are therefore up to around index 60.
}
    \label{fig:buffer_fill_level}
\end{figure}

Since the PAS and the FAS are connected via a limited buffer, the buffer acts as a bottleneck that constrains the material flow.
The buffer fill level increases over time (Fig. \ref{fig:buffer_fill_level}). This is intuitive, since the agent causes products to be produced and the PAS has a higher throughput than the FAS. This effect becomes particularly significant later in the planning period.

\subsubsection{Buffer fill level for the specific products}
\begin{figure}[H]
    \centering
    \includegraphics[width=1\linewidth]{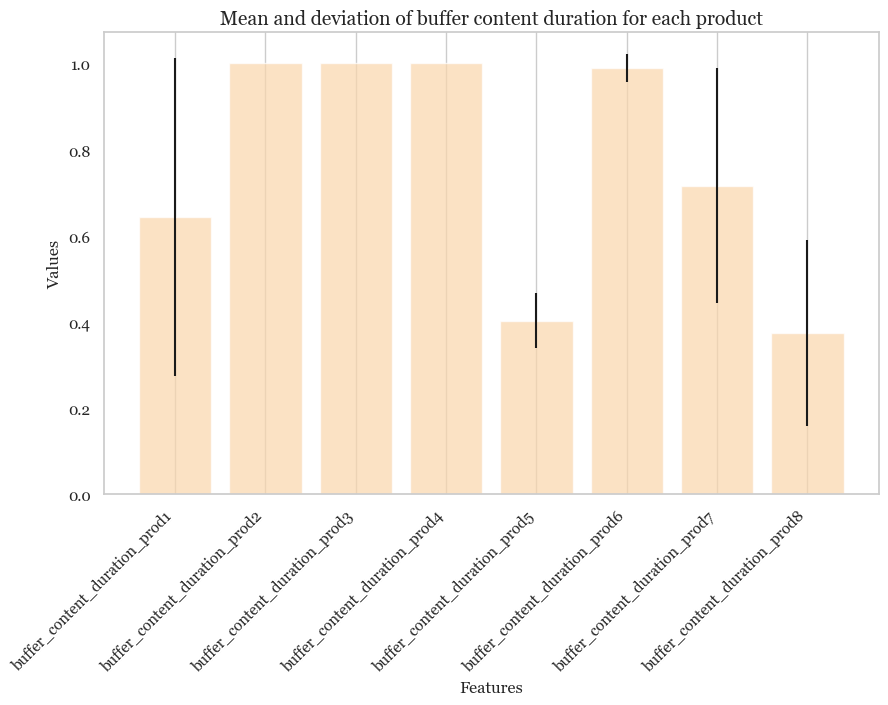}
    \caption{Mean buffer content duration for products. The vertical lines show standard deviations.}
    \label{fig:buffer_for_aggs}
\end{figure}

To gain knowledge about specific products, the role of the buffer has to be examined for these product types. The following plot shows the mean buffer content for all products over the whole trajectory (Fig. \ref{fig:buffer_for_aggs}).

Utilizing this feature, we can interpret the production of these products. 
The products that were \textit{not} produced (2, 3, 4, 6) had a mainly full mean buffer. Since the buffer satisfies the demand, products with a full buffer were not produced. The product types that were produced had a lower mean buffer and more deviation overall. In fact, product 5 – produced the most – had the lowest mean buffer. Here, it is important to highlight that the agent tries to keep a sufficient level in buffer for all product types to maintain the safety margin and avoid criticality. Therefore, it is plausible to assume that those products with a low buffer fill level are produced more often. Vice versa, if the \texttt{buffer\_content\_duration} of a given product is high, it is more unlikely that this product will be produced. However, this only accounts for criticality. When considering setup efforts the effect can be reversed. The probability of producing an product can increase although the \texttt{buffer\_content\_duration} of this product is already high to minimize setup efforts. Thus, the criticality and setup efforts hypotheses may be of relevance simultaneously and their effects can overlap.

\subsubsection{Demand}
Demand is encoded in two features, \texttt{next\_24h\_demand} and \texttt{end\_of\_planning\_period\_demand}. The adjoining table summarizes these:

\begin{table}[ht!]
    \footnotesize
    \centering
    \caption{The mean demand for the products. * = produced}
    \label{t:demand_agg}
    \begin{tabularx}{\textwidth}{lXX}
        \toprule
        \textbf{Product} & \textbf{next\_24h\_demand} & \textbf{end\_of\_planning\_period\_demand} \\
        \midrule
        1* & 0.002636 & 0.068180 \\
        2 & 0.000000 & 0.000250 \\
        3 & 0.000000 & 0.000000 \\
        4 & 0.000000 & 0.000000 \\
        5* & 0.063485 & 1.000000 \\
        6 & 0.000000 & 0.000000 \\
        7* & 0.013801 & 0.330981 \\
        8* & 0.041675 & 0.857493 \\
        \bottomrule
    \end{tabularx}
\end{table}
\FloatBarrier

It is important to note that both demand types, \texttt{next\_24h\_demand} and \texttt{end\_of\_planning\_period\_demand}, show the demand left after considering the amount of demand the buffer can satisfy. Therefore, a logical conclusion is that if the buffer for an product increases the demand should decrease.
Since the 24h demand is more urgent in a manufacturing environment, those products with a higher \texttt{next\_24h\_demand} should be more likely to be produced. In fact, we see in Table \ref{t:demand_agg} that only those products with a 24h demand were produced at all. Product 5 with the highest mean demand (both types) was produced the most. 

The conclusion to be derived from this is that net demand (after deducting the buffer) of an product is positively associated with its production. However, if no product is critical, the agent can continue to produce an product with low demand to minimize setup efforts. Additionally, demand of other products can also influence if an product with high demand is produced. If the demand of another product is higher, a high demand of the product being produced in the instance can make it more unlikely that it is being produced again, because the criticality of another product is higher and the agent has to switch to producing this product.

\subsubsection{Criticality}
\label{criticality_chapter}
Criticality \( C \) is defined as:

\[
C = \frac{\texttt{next\_24h\_demand}}{\texttt{buffer\_content\_duration}},\]

where \(\texttt{buffer\_content\_duration} > 0\). Increasing criticality is linked to increasing demand and decreasing buffer content of an product. Criticality is implicitly present in the data, but not explicitly as a feature. We assume that the NN was able to learn this non-linear relationship \shortcite{bengiononlinear}.

\subsection{Generating hypotheses}
The hypotheses to be derived using domain knowledge (RQ 1.1), descriptive statistics and the objectives of the agent can be subsumed under the two parts of the agent's reward function.

\textbf{Criticality}:
High criticality is characterized by high net demand and low buffer content, which is a state the agent tries to avoid, because it increases risk of idle times.
Therefore, a high net demand and/or low \texttt{buffer\_content\_duration} of one of the produced products in the data should be positively associated with production in the xAI methods and vice versa. However, if the demand of the\textit{ successing }product is higher and/or the buffer is lower, the agent should switch to that product and \texttt{last\_prod\_type} = 1, a higher demand and/or lower\textit{ }buffer of the current product will speak \textit{against} it being produced again.
In Table \ref{tab:criticality_analysis}, the criticality for product 5 is zero (the demand is zero and the buffer is halfway full (0.522), but the criticality for product 8 is greater than 1 (there is demand and the buffer content is low). Therefore, the agent switches from producing product 5 to product 8 in the second action.

\textbf{Setup efforts}:
The agent tries to minimize setup efforts, which is indicated by the \texttt{last\_prod\_type} dummy variable. If an product was produced already (\texttt{last\_prod\_type} = 1), it should increase the probability that the same product will be produced again. This however can lead to a trend that speaks against the hypothesis implied by criticality: The probability of producing an product may increase, although the buffer of this product is high and the demand is low. This can happen, when no other product is critical; therefore, the agent sticks to the production of one product to minimize setup efforts.
If we find the reversed trend and \texttt{last\_prod\_type} = 1 decreases the probability that the product is produced again, this should correspond with high criticality – the first objective of the agent – for a different product that is then produced in the next instance.

\begin{table}[ht!]
    \footnotesize
    \centering
    \caption{Criticality Analysis}
    \label{tab:criticality_analysis}
    \begin{tabularx}{\textwidth}{lXr}
        \toprule
        \textbf{Product} & \textbf{Metric} & \textbf{Value} \\
        \midrule
        Product 5 & Criticality & 0 \\
                    & 24h\_demand & 0 \\
                    & buffer\_duration & 0.522 \\
                    & last\_prod\_is\_prod\_5 & 1.0 \\
        \midrule
        \textbf{Product 8} & \textbf{Criticality} & \textbf{1.098} \\
                    & \textbf{24h\_demand} & \textbf{0.191} \\
                    & \textbf{buffer\_duration} & \textbf{0.174} \\
                    & \textbf{last\_prod\_is\_prod\_5} & \textbf{0} \\
        \midrule
        Product 8 & Criticality & 0.875 \\
                    & 24h\_demand & 0.175 \\
                    & buffer\_duration & 0.200 \\
                    & last\_prod\_is\_prod\_8 & 1.0 \\
        \midrule
        Product 8 & Criticality & 0.716 \\
                    & 24h\_demand & 0.159 \\
                    & buffer\_duration & 0.222 \\
                    & last\_prod\_is\_prod\_8 & 1.0 \\

        \bottomrule
    \end{tabularx}
\end{table}
\FloatBarrier

\begin{table}[ht!]
    \footnotesize
    \centering
    \caption{Setup-efforts Analysis}
    \label{tab:setup_efforts_analysis}
    \begin{tabularx}{\textwidth}{lXr}
        \toprule
        \textbf{Product} & \textbf{Metric} & \textbf{Value} \\
        \midrule
        Product 5 & Criticality & 0.887 \\
                    & 24h\_demand & 0.268 \\
                    & buffer\_duration & 0.302 \\
                    & \textbf{last\_prod\_is\_prod\_5} & \textbf{1.0} \\
        \midrule
        Product 5 & Criticality & 0.784 \\
                    & 24h\_demand & 0.243 \\
                    & buffer\_duration & 0.310 \\
                    & \textbf{last\_prod\_is\_prod\_5} & \textbf{1.0} \\
        \midrule
        Product 5 & Criticality & 0.735 \\
                    & 24h\_demand & 0.228 \\
                    & buffer\_duration & 0.310 \\
                    & \textbf{last\_prod\_is\_prod\_5} & \textbf{1.0} \\
        \midrule
        Product 5 & Criticality & 0.732 \\
                    & 24h\_demand & 0.227 \\
                    & buffer\_duration & 0.310 \\
                    & \textbf{last\_prod\_is\_prod\_5} & \textbf{1.0} \\
        \bottomrule
    \end{tabularx}
\end{table}
\FloatBarrier

These two tables illustrate both objectives of the agent with examples from production. 
    In Table \ref{tab:criticality_analysis}, we can see that product 8 is critical (in bold), because there is demand for this product, but the buffer content is low. The agent then switches from producing product 5 with zero criticality to production of this product. 
    In Table \ref{tab:setup_efforts_analysis}, setup efforts are minimized by sticking to the production of the same product, which is indicated by the last product type always being the same (in bold).

As both tables suggest, the hypotheses need to be tested with the produced products, taking all features into account (RQ 1.1) and embedding the complete production scenario. We now look at the top ten most important variables and products for both xAI methods. Then, we compare both methods regarding their performance and interpretability. We also apply the hypotheses to the products not produced (see Chapter \ref{Appendix_Notproduced_Aggs}).

\subsection{Applying hypotheses}
\subsubsection{Product 1}
Product 1 was produced the third most (Fig. \ref{fig:order}). In the order of production, it predecessor and successor product was product 5 which was the most produced product.

\begin{figure}[H]
    \centering
    \includegraphics[width=0.75\linewidth]{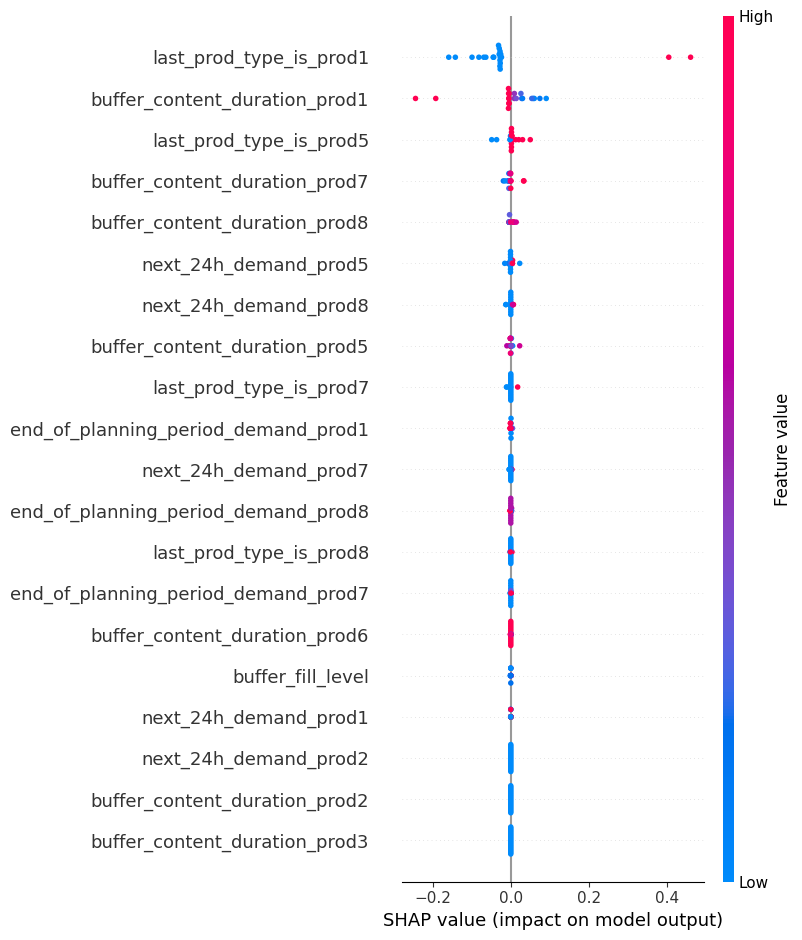}
    \caption{SHAP summary plot for action 0. 
Each point represents the local feature attribution value (Shapley value for feature and instance).
Blue color indicates a low feature value, for binary variables this is 0, red indicates high feature values, for binary variables this is 1. A positive SHAP value is positively associated with the action, a negative SHAP value is negatively associated with the action. The features are displayed based on importance on average with decreasing importance from top to bottom.}
    \label{fig:SHAP_Action0}
\end{figure}

Regarding DeepSHAP, we can see that if the last product produced was already product 1, it is more likely that this product will be produced again (Fig. {\ref{fig:SHAP_Action0}}) and vice versa. This aligns with the setup efforts hypothesis. Moreover, a lower buffer content is positively associated with production of the product and vice versa, which matches the criticality hypothesis.
If the last product type was product 5, product 1 is more likely to be produced. This aligns with product 1 only being produced following product 5 in the course of production. The other variables have SHAP values around zero. 

Regarding Input X Gradient (Table \ref{tab:variables}), it can be seen that increasing \texttt{buffer\_content\_duration\_prod1} mitigates the production of product 1. This can be taken one step further by observing that a higher buffer content for product 5 supports the production of product 1. This is in line with the criticality hypothesis, because product 5 is less critical based on a full buffer. In the result, product 1 is produced more. If the last product which was produced is product 1 as well, it has a positive impact on the renewed production of this product. Here the setup-efforts hypothesis is confirmed and matches the data that shows that product 1 was produced continuously without a machine retrofitting. If the \texttt{end\_of\_planning\_period\_demand} of product 5 or 8 were increasing, product 1 had slightly less attribution which matches the criticality hypothesis. The demand at the end of the planning period for both product 5 and 8 favoured less production of product 1, but this effect was really low.
Both hypotheses categories are thus supported by Input X Gradient for product 1.

\subsubsection{Product 5}
Product 5 was produced the most (Fig. \ref{fig:order}). Its production was interrupted by the products 8 and 1, so the machines had to be retrofitted.

\begin{figure}[H]
    \centering
    \includegraphics[width=0.85\linewidth]{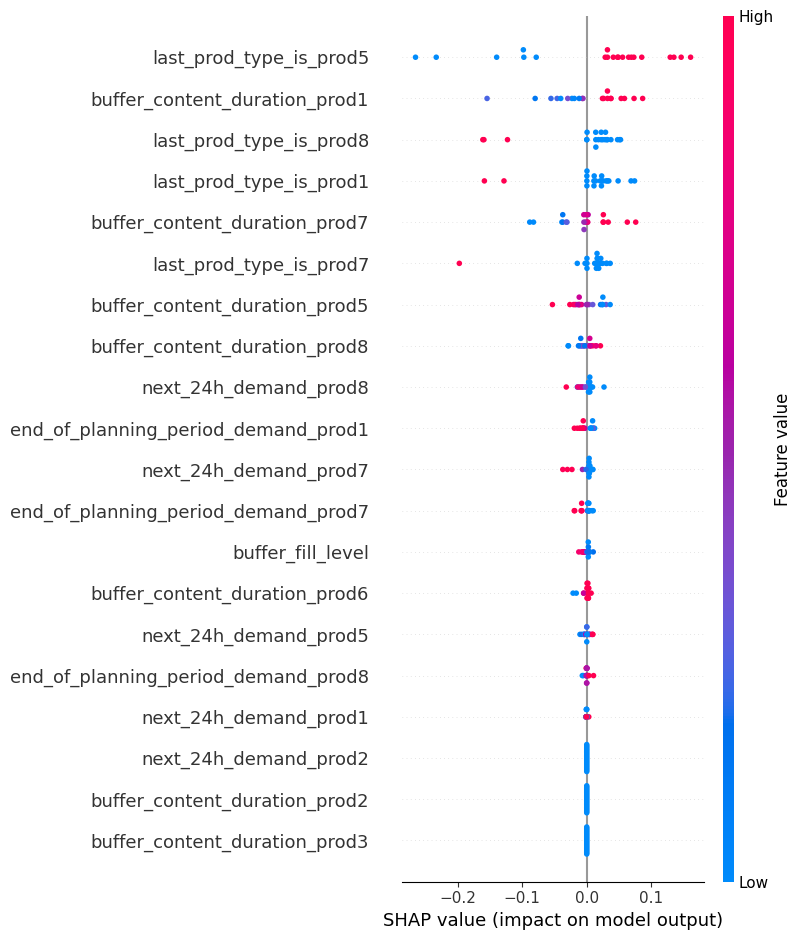}
    \caption{SHAP summary plot for action 4. 
Each point represents the local feature attribution value (Shapley value for feature and instance).
Blue color indicates a low feature value, for binary variables this is 0, red indicates high feature values, for binary variables this is 1. A positive SHAP value is positively associated with the action, a negative SHAP value is negatively associated with the action. The features are displayed based on importance on average with decreasing importance from top to bottom.}
    \label{fig:SHAP_Action4}
\end{figure}

Regarding DeepSHAP, we can see that if product 5 was produced last, it is more likely to be produced again (Fig. \ref{fig:SHAP_Action4}), which is in line with the setup efforts hypothesis.
A full buffer content of product 1 is positively associated with production of product 5, while an empty buffer of product 1 speaks against production of product 5. This pattern supports the criticality hypothesis, because if product 1 becomes critical (e.g., due to a low buffer content), production of product 5 should be less likely and vice versa, which is what we find here.
In the order of production, product 8 was produced twice following product 5, but product 5 was not produced after product 8. Therefore, it makes sense that the production of product 5 is more likely when product 8 was not previously produced. However, it should be noted that there are two outliers indicating the opposite direction of effect.
Product 1 disrupted production of product 5 once. The pattern of the variable \texttt{last\_prod\_type\_is\_prod1} is similar to the one of product 8. Most of the time, product 5 was not produced after product 1 and as we can see in the SHAP plot, production of product 5 is associated with the previous product not being product 1. However, there are two outliers again.
Fuller buffer content of product 7 speaks for production of product 5 and vice versa. This is again in line with the criticality hypothesis.
Next, we see that if the last product produced was product 7, this speaks against product 5 being produced next. Here, we have to take into account that product 5 was produced the most, product 7 was produced the least. There is only one case in the order of production, where product 5 was produced after product 7; however; in all other cases this was not true. Therefore, the pattern aligns with the course of production.
The other findings in the SHAP plot, all support the criticality hypothesis:
A low buffer content of product 5 is positively associated with production of product 5.
A higher buffer content of product 8 speaks for production of product 5, while a higher 24h-demand of product 8 is negatively associated with production of product 5. 
Higher end of planning period demand for product 1 also speak against production of product 5. 

The following table shows the attribution values we received with the Input X Gradient method:

\begin{table}[ht!]
    \footnotesize
    \centering
    \caption{Top 10 Variables for Input X Gradient and Product 5.}
    \label{tab:top_variables_target4}
    \begin{tabularx}{\textwidth}{lXr}
        \toprule
        \textbf{Rank} & \textbf{Variable} & \textbf{Value} \\
        \midrule
        1  & buffer\_content\_duration\_prod5 & -0.17 \\
        2  & buffer\_content\_duration\_prod8 & 0.12 \\
        3  & last\_prod\_type\_is\_prod5 & 0.10 \\
        4  & end\_of\_planning\_period\_demand\_prod5 & 0.08 \\
        5  & buffer\_content\_duration\_prod1 & 0.07 \\
        6  & buffer\_content\_duration\_prod6 & -0.06 \\
        7  & next\_24h\_demand\_prod8 & -0.05 \\
        8  & end\_of\_planning\_period\_demand\_prod8 & -0.04 \\
        9  & buffer\_content\_duration\_prod3 & 0.03 \\
        10 & buffer\_content\_duration\_prod4 & -0.02 \\
        \bottomrule
    \end{tabularx}
\end{table}
\FloatBarrier

If the buffer content was high for product 5, the probability that it was produced again sunk. In addition, if the buffer for product 8 was high, product 5 was produced more often. A plausible conclusion could be that the buffer content was sufficient to satisfy the demand of product 8, thus alleviating criticality. Unsurprisingly, if the last product was product 5 it was more likely that it was produced again (\texttt{last\_prod\_type\_is\_prod5}). The setup efforts hypothesis is thus supported here. The \texttt{end\_of\_planning\_period\_demand\_prod5} also had a positive influence. The machine had to be retrofitted two times for products 1 and 8, high buffer content for these products amongst other effects triggers product 5's production as well. A full buffer signals that product 5 can be produced again (lower criticality of products 1 and 8). Also slightly supporting the criticality hypothesis, a higher demand for product 8 in 24h had a negative attribution for product 5. 

Both hypotheses are supported thus by Input x Gradient for product 5.

\subsubsection{Product 7}
Product 7 was produced the least. The predecessor of product 7 was
product 8, the successor product 5. 

\begin{figure}[H]
    \centering
    \includegraphics[width=0.75\linewidth]{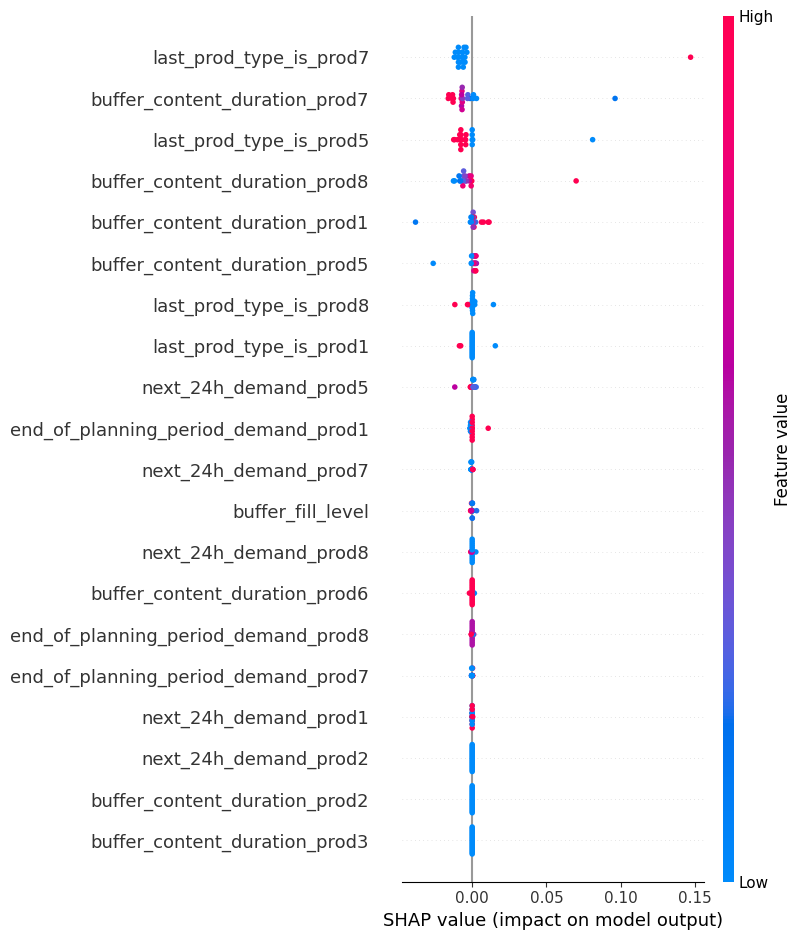}
    \caption{SHAP summary plot for action 6. 
Each point represents the local feature attribution value (Shapley value for feature and instance).
Blue color indicates a low feature value, for binary variables this is 0, red indicates high feature values, for binary variables this is 1. A positive SHAP value is positively associated with the action, a negative SHAP value is negatively associated with the action. The features are displayed based on importance on average with decreasing importance from top to bottom.}
    \label{fig:SHAP_Action6}
\end{figure}

Regarding SHAP (see Fig. \ref{fig:SHAP_Action6}), we can see that if product 7 was not produced last, production of this product is less likely, which is in line with the setup efforts hypothesis. There is only one individual SHAP point indicating the opposite direction of this hypothesis (i.e., that production of product 7 is positively associated with it being produced again). This makes sense, because descriptively this product was produced only a couple of times, too few cases to create a visual trend in
the plot.
Fuller buffer content of product 7 speaks against its production, which is in line with the criticality hypothesis; however, the SHAP values for the opposite direction of this hypothesis (i.e. less buffer content is associated with more production) are close to zero. Again, there is only one individual point indicating the latter, which might be tied to product 7 being rarely produced. If it was barely ever produced, it is intuitive that there will be only few cases indicating such a direction of effect.
Product 7 was never produced directly after product 5, which aligns with the plot where we can see that if the last product produced was 5, production of product 7 is less likely.
Lower buffer contents of product 8 is negatively associated with production of product 7, in alignment with the criticality hypothesis. However, the SHAP values for the opposite direction of this hypothesis are close to zero.
Fuller buffer contents of product 1 and 5 are positively associated with production
of product 7, in line with the criticality hypothesis. However, again the SHAP values are close to zero.

Regarding Input x Gradient (Table \ref{tab:top_variables_target6}), the attributions were generally low. A higher \texttt{buffer\_content\_duration\_prod\_8} compliments the production of product 7, whereas a higher product 7 buffer speaks against its production. This supports the criticality aspect (see above). Interestingly, the generic buffer fill level has a negative influence on the production of product 7. This may be because of its rare occurrence. In summary Input x Gradient supports the criticality hypothesis.

\subsubsection{Product 8}
Product 8 was produced the second most.

\begin{figure}[H]
    \centering
    \includegraphics[width=0.75\linewidth]{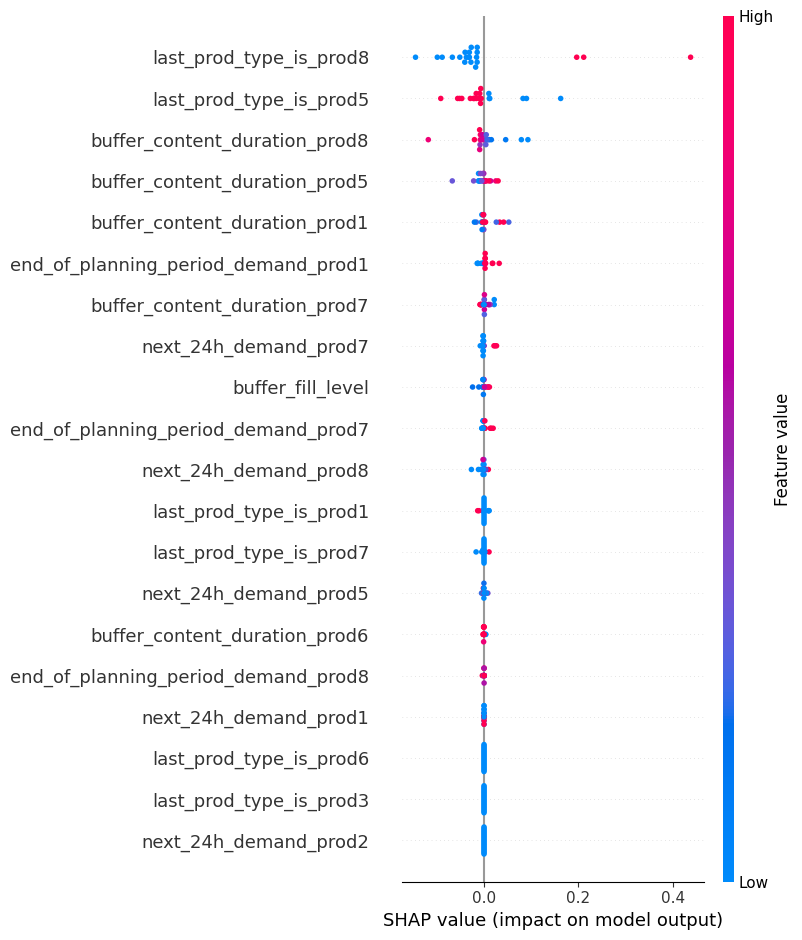}
    \caption{SHAP summary plot for action 7. 
Each point represents the local feature attribution value (Shapley value for feature and instance).
Blue color indicates a low feature value, for binary variables this is 0, red indicates high feature values, for binary variables this is 1. A positive SHAP value is positively associated with the action, a negative SHAP value is negatively associated with the action. The features are displayed based on importance on average with decreasing importance from top to bottom.}
    \label{fig:SHAP_Action7}
\end{figure}

Regarding DeepSHAP, we can see that if product 8 was not produced last, it is less likely to be produced again (Fig. \ref{fig:SHAP_Action7}). This is generally in line with the logic of the setup efforts hypothesis, though it is the reversed pattern (e.g., the product \textit{not} being produced last is associated with it \textit{not} being produced again). Interestingly, for the other direction of effect (e.g., the product being produced last, is associated with it being produced again) there are only few individual SHAP points supporting this trend. This could be related to product 8 batches only being produced thirteen total times, while product 5 was produced more than five times more often. Therefore, it is plausible that the effect is smaller for product 8 than it is for product 5.
Previous production of product 5 is negatively associated with production of product 8, which is in line with production order, because product 5 was produced in the biggest batches without interruption and there are only two cases in total where product 8 was produced after product 5. 
Less buffer content duration of product 8 is positively associated with production of product 8, which is in line with criticality hypothesis. Further, a fuller buffer of product 5 (e.g., criticality is low) speaks for production of product 8. Though the SHAP values are small, this matches the criticality hypothesis.
The pattern for buffer content duration of product 1 is ambiguous, but here the SHAP values are close to zero as well
A higher end of planning period demand of product 1 speaks for production of product 8, which might seem unintuitive. Here, it is important to differentiate between the two types of demand in the domain. For criticality, the 24h-demand plays the most important role. This means that the end of planning period demand of an product may be high, but if another product is \textit{more} critical in the next 24h it is going to be produced instead. In this particular case, the variable of buffer content duration of product 8 was higher up in the order of variable importance than end of planning period demand of product 1. This might indicate that product 8 was more critical due to an empty buffer, which is why it was produced; even though, end of planing period demand of product 1 was high.
The other variables have SHAP values around zero. 

Regarding Input x Gradient (Table \ref{tab:top_variables_target7}), it can be derived that a higher buffer of product 8 was not supporting further production of product 8 (supporting the criticality hypothesis). If the buffer for product 5 increased in contrast, this supported the production of product 8. Here, the inverse relationship of the criticality hypothesis is visible: If the buffer for product 8 was high, the production of it sunk. If the buffer for product 5 was high, the production of product increased. This can also be seen in the light of the predecessor/successor relation: Product 8 had been uninterruptedly produced for a longer period of time (product 8 was produced after product 5 two times (four times and nine times respectively). The demand of the end of the planning period of product 8 slightly increased the chances of producing product 8. The criticality hypothesis can thus be confirmed, but not with total confidence. The attribution values were too low. Regarding the setup efforts hypothesis, Input x Gradient does not fully support this hypotheses for product 8. 

\subsection{Verifying hypotheses}
Regarding DeepSHAP, we see that the hypotheses of avoiding criticality and minimizing setup efforts hold true both in the patterns for the variables directly related to the produced products and also for the variables concerning the other products.
Input x Gradient supports both hypotheses as well for product 1 and 5. Product 8's attribution values were too low to draw a viable conclusion. 

\subsubsection{Variables related to the produced products}
In DeepSHAP, we can see that if the last product produced was already product 1, product 5, or product 8, it is more likely that these products will be produced again and vice versa, which aligns with minimizing setup times.
A lower buffer content of product 1, product 5, or product 8, is positively associated with production of these products respectively and vice versa, which is in line with the criticality hypothesis.
Due to it rarely being produced, we can also verify the hypotheses for product 7, but only in one direction: If product 7 was not produced last, production of this product is less likely. Fuller buffer content of product 7 speaks against its production.

Input x Gradient also confirms the setup time hypothesis with the feature \texttt{last\_prod\_type\_is\_prodX} for product 1 and 5 but not for product 7 and 8.
Increasing \texttt{buffer\_content\_duration\_prodX} for all four products confirms the criticality aspect as well. A fuller buffer signals lower criticality. The buffer is full and thus it is not critical to produce the product. 

\subsubsection{Variables for other products}
In DeepSHAP, a fuller buffer content of product 1, product 7 or product 8 speak for production of product 5 and vice versa, while a higher demand of product 8 or product 1 speaks against production of product 5. This is in line with the criticality assumption. If the buffer of these products is low or the demand is high and they become critical, product 5 should be less likely to be produced to avoid idle times. However, if they are not critical, the agent should stick to production of product 5 to minimize setup efforts, which is what we see in the plots for product 5. 
This pattern holds true for the other classes as well, e.g., a fuller buffer content of product 5 speaks for production of product 8. Lower buffer contents of product 8 is negatively associated with production of product 7.

In the case of Input x Gradient, a full buffer for product 5 supported the production of product 1 (because the buffer of product 5 was high, yielding low criticality for this product). If the demand of product 5 or 8 were increasing, the influence of product 1 decreased, because the criticality of product 5 and 8 increased. The demand at the end of the planning period for both product 5 and 8 favoured less production of product 1, but this effect was low.
For product 5, other features also played an important role. A high buffer content for product 8 supported the production of product 5, because the criticality of product 8 sunk. This holds true for the products 1 and 6 as well. Also slightly supporting criticality, a higher demand for product 8 in 24h had a negative influence on the production of product 5. 
For product 8, fuller buffer content for product 5 supported the production of product 8. Again, because the criticality of product 5 sunk. If the last product was product 5, the probability that product 8 was produced sunk slightly. Note that product 8 was not produced perpetually and was interrupted by product 7. It also was produced after 33 products of 5 were produced, so the machine was retrofitted at the end.
For product 7 - the least produced one - we see the criticality hypothesis confirmed because of a higher buffer content of its predecessor product 8. Analogous, a higher buffer for the buffer of 7 itself lowers the probability that it is produced again. 

In summary, the overall patterns in the DeepSHAP plots match our hypotheses. Considering Input x Gradient, in the case of product 1 the setup effort hypothesis can be confirmed because there it uninterrupted production. The criticality hypothesis is confirmed as well.
For product 5, both hypotheses can be confirmed again. The reduction of retrofitting can be confirmed with \texttt{last\_prod\_type\_is\_prod5}.
Product 8 only slightly favours the criticality hypothesis. There is a negative attribution for the \texttt{buffer\_content\_duration\_prod8} which favors to the criticality aspect (compare above). Product 7 favours the criticality hypothesis. Its attribution values were low as well.

\subsubsection{Robustness checking}
In order to ensure transferability and robustness of our approach, we tested our hypotheses on a different week of production with a different data constellation (only products 4 and 5 were produced). Here, our hypotheses match the DeepSHAP results as well (see \ref{Appendix_SHAP}). For example, if product 5 (\ref{fig:agg5-week42}) was produced last, it is more likely to be produced again (setup efforts hypothesis) and if the buffer content of product 5 is lower or the demand is higher, production of product 5 is more likely (criticality hypothesis).
Input x Gradient does not support these hypotheses, because the attributions yield contrary results because of only two actions. The buffer content duration for product 7 and the buffer content duration for product 2 give conflicting attribution values for products 4 and 5 (inverse correlations). On the later attributions we see the setup efforts hypothesis confirmed for product 4 (\textit{last\_pr\_type\_is\_pr4}) is positive).

\subsubsection{Falsification of explanations}
We were able to validate the xAI explanations by comparing them to the hypotheses we generated in advance based on domain knowledge. Now, we can present the hypotheses as interpretations of the agent's behavior to the stakeholders, instead of having non-experts in AI inspect dozens of plots and attribution values. The production planners (or management) can then in return give feedback on the plausibility of these explanations.

\subsection{Comparing DeepSHAP and Input x Gradient}
DeepSHAP matched the hypotheses more often in our use case. Using SHAP values, different effects for higher and lower feature values can be inspected, which is not the case for Input x Gradient and very beneficial to gain a deeper understanding in the decision process of the agent. Also, SHAP offers a range of different visualizations that can facilitate understanding. However, in DeepSHAP there are single outliers and unreasonable data points as previously discussed in \ref{SHAP} and this method might lead to several hurdles along the way when confronted with variables that are highly dependent. Furthermore, a background dataset has to be chosen; therefore, DeepSHAP is only applicable for use cases that create larger datasets. We conclude that DeepSHAP performs better for classes with many cases (e.g., product 5) and worse for classes with few cases (e.g., product 7). It should also be considered that the concept of Shapley values may be harder to communicate.

Input x Gradient favours the hypotheses less often than DeepSHAP. The interpretation of the explanations is not visibly attractive as in the case of DeepSHAP. Attribution plots can be generated to compare the attribution values for different products and features. It can not be directly compared because the single observations are not visible in the plot like in the SHAP plot. Different gradient attribution approaches are more often used in the context of convolutional NNs. It is harder to use them extensively for tabular data. In the case of the most produced products, attribution values can be computed efficiently and support our findings. In case of the other less produced products like product 7, the attribution values are small (see \ref{Appendix_InputxGradient}) and can thus not be interpreted meaningfully. 

Overall, SHAP was more suitable for the given data (RQ 1.2). Generally, it is desirable to be able to interpret all actions of the agent. For Input x Gradient, this was not possible. The dataset was limited to specific actions and several features showed slight attribution for non-produced products. In a practical setting, this may be hard to communicate. SHAP might be the way to go here, because it may provide better explanations. Another approach may be to gather more data, if possible. This would enhance both methods. SHAP's plotting functionality works out of the box and may assist better in finding interpretations. 

\subsection{Adapted xAI workflow}
Using the results and insights we gained by generating explanations for the DRL agent in our use case based on the workflow of \citeA{tchuente2024methodological}, we now adapt said workflow to make it practically feasible for using xAI in the context of production planning (RQ 2.1). We could apply the greyed out phases unfetteredly in this special use case. To apply the whole workflow for explaining DRL agents in a production scheduling context, we made several adjustments.

\begin{figure}[H]
    \centering
    \includegraphics[width=0.8\linewidth]{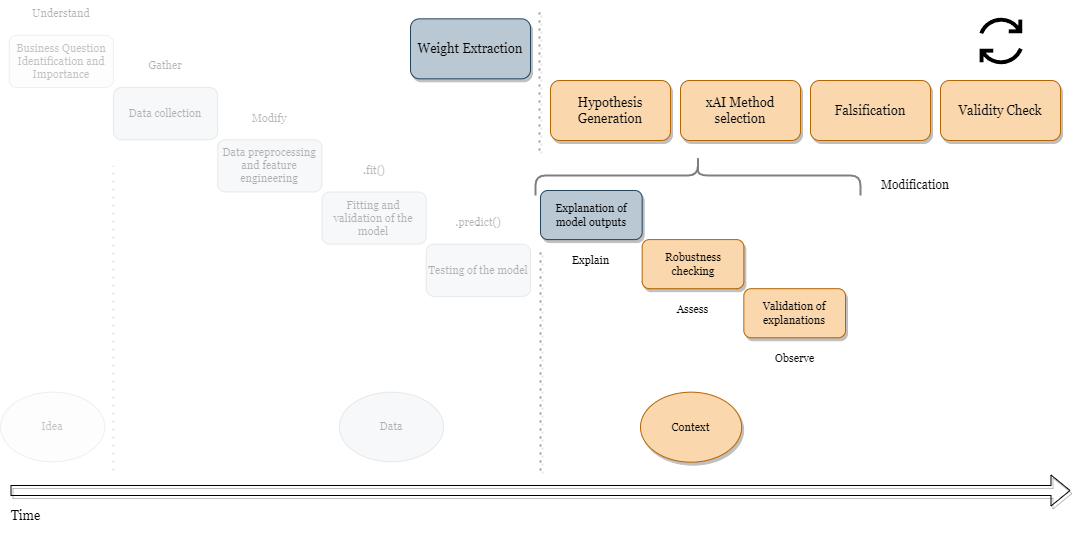}
    \caption{Our proposed framework is rooted in \captionciteA{tchuente2024methodological}, but introduces new phases.}
    \label{fig:tchuente_modified}
\end{figure}

\subsubsection{\textbf{Weight Extraction}:}
The first novel phase consists of extracting the weights of the network and gathering the available data. This is particularly important, because in RL a direct mapping between states and actions is oftentimes not possible and thus the function is approximated, for example by using NNs. By extracting the weights, we have access to the learned policy.

    The approach must be adapted depending on the data available:

    \begin{itemize}
        \item If only the raw data of the agent's decision is available and there is no access to model weights or other information regarding the network, it is feasible to train a surrogate model on the data frame. For example, a random forest classifier can be used to model the agent's behaviour. This classifier can then serve as an input for various xAI frameworks. The information gain here is limited, because the agent used a different model.

        \item Generating new weights is also a feasible approach. When only a data frame is available and the net dimensions are clear, one can mimic the network structure and create a new network with the same proportions (\textit{fake network}). Then, use inference on the data frame and describe its new weights with the xAI frameworks. The information gain here is better than in the latter approach although still noisy. In the network, the structure is the same but the weights are different. In the context of post-hoc model agnostic explainers, using different weights is not optimal. This is because ideally, we want to explain the models "ideas" during the training phase. 

        \item When the data, weights and network structure (not the network itself) are at hand, these can be used to generate the explanations. In our use case, this was achieved by the export\_model function of the Ray framework, which was utilized to train the agent \shortcite{moritz2018ray}. If the network has been implemented in PyTorch \shortcite{paszke2019pytorchimperativestylehighperformance}, extracting the weights is not needed. For our use case, we mimicked the network structure in PyTorch and copied the weights from the Ray model. You may extract certain layers to explain specific structures of the net. After this, we loaded the network with the classical \textit{torch.load} command.

        \item If the real network (e.g. not a Ray version), the weights and the raw data is available, you may directly apply the xAI methods on the net. 
    \end{itemize}

This step ends the data phase.
    
\subsubsection{\textbf{Hypotheses generation}:}
After we extracted the weights, we can start by formulating hypotheses about the agent's behaviour. For this, we consider causal presumptions based on domain-knowledge and the reward function, as well as inspecting descriptive statistics. This approach works for xRL as well as for other xAI; for cases outside of DRL, the knowledge of the reward function has to be exchanged with other model specific assumptions or be left out.
Hypotheses can be formulated using positive or negative relationships between the variables. It is also possible to construct relationships between hypotheses.
Note that if the reward function is already constructed in an interpretable way \textit{before} training (e.g., by reward decomposition), this will simplify this approach. However, for post-hoc explainabilty, influencing stages before the model was employed is not possible.

\subsubsection{\textbf{xAI method selection:}}
Once the hypotheses stand, we can choose suitable xAI frameworks and methods identified in the information phase. Randomness may arise in the split of training and test data, the explanation method itself or in the training of the network. Setting seeds in this context is a safe way to achieve fixed explanations.
It is desirable to apply frameworks which cover a wide range of xAI methods. Here, one might apply a model-specific and a model-agnostic xAI method or use local and global explainers. For this step, a cooperation with data scientists might be beneficial. We give an overview of recommendations; however, these are not comprehensive and an analysis on the requirements, constrains, and characteristics of the use case must be performed to find the right explainer:
    \begin{enumerate}
        \item Model-specific explainers might be used when debugging is the main goal or when computational costs are high and can be lowered using model-speciﬁc information.
        \item Model-agnostic methods can be utilized when model-specific methods are constructed too narrowly and cannot be applied to complex real-world scenarios.
        \item Local explanations may be incorporated when only single decisions are to be explained.
        \item Global explainers should be used when the behavior of the (entire) model is of interest.

    \end{enumerate}
 To enhance robustness, it is possible to fit the explainer on different configurations of the network and to use cross validation in the same step \cite{brownecrossval}. After capturing the explanation values, it is important to check for outliers in the generated list of attributions. Also, the attributions may be aggregated using the mean or median. A confidence interval may be fitted to further support the findings \shortcite{napolitano2023learning, pmlr-v238-neuhof24a}. \citeA{napolitano2023learning} for example develop Interval FastSHAP based on Coalition Interval Games and IntervalSHAP. They show that the prediciton quality can be increased by using more predictors.

\subsubsection{\textbf{Falsification}:}
 Now, we apply the xAI methods and interpret the findings using our hypotheses to verify or falsify the explanations: Do the hypotheses serve as an interpretation of the xAI explanations? If so, the xAI results align with domain knowledge and the hypotheses can be communicated as interpretations of the agent's behavior (green light in Figure \ref{fig:validitycheck}). If explanations and hypotheses do not match, visit the section validity check.

\subsubsection{\textbf{Validity Check}:}
The goal here is to inspect inconsistencies between the xAI methods, the hypotheses and the broader context (yellow and red light in Figure \ref{fig:validitycheck}). Investigate where and to what degree there are deviations between the hypotheses and the xAI explanations, then compare both with domain-knowledge and descriptive statics (e.g., the broader context).
    \begin{enumerate}
        \item Do the xAI results match the broader context? In this case, you may have missed crucial elements in the hypotheses generation. Improve your hypotheses and retest them, preferably on a new data set.
        \item Do the xAI results still speak against domain-knowledge? If so, you need to double-check the validity of the chosen xAI methods (assumptions, prerequisites, plausibility for the use case) and possibly apply the hypotheses to other xAI methods. If the mismatch between hypotheses and xAI results persists, you may have detected issues in the underlying AI-model and need to consult developers.
        \end{enumerate}

To fully cover for the non-deterministic nature of the explanations, we deliver an extension of the proposed framework which highlights how the Validity Check can be characterized:

\begin{figure}[H]
    \centering
    \includegraphics[width=0.8\linewidth]{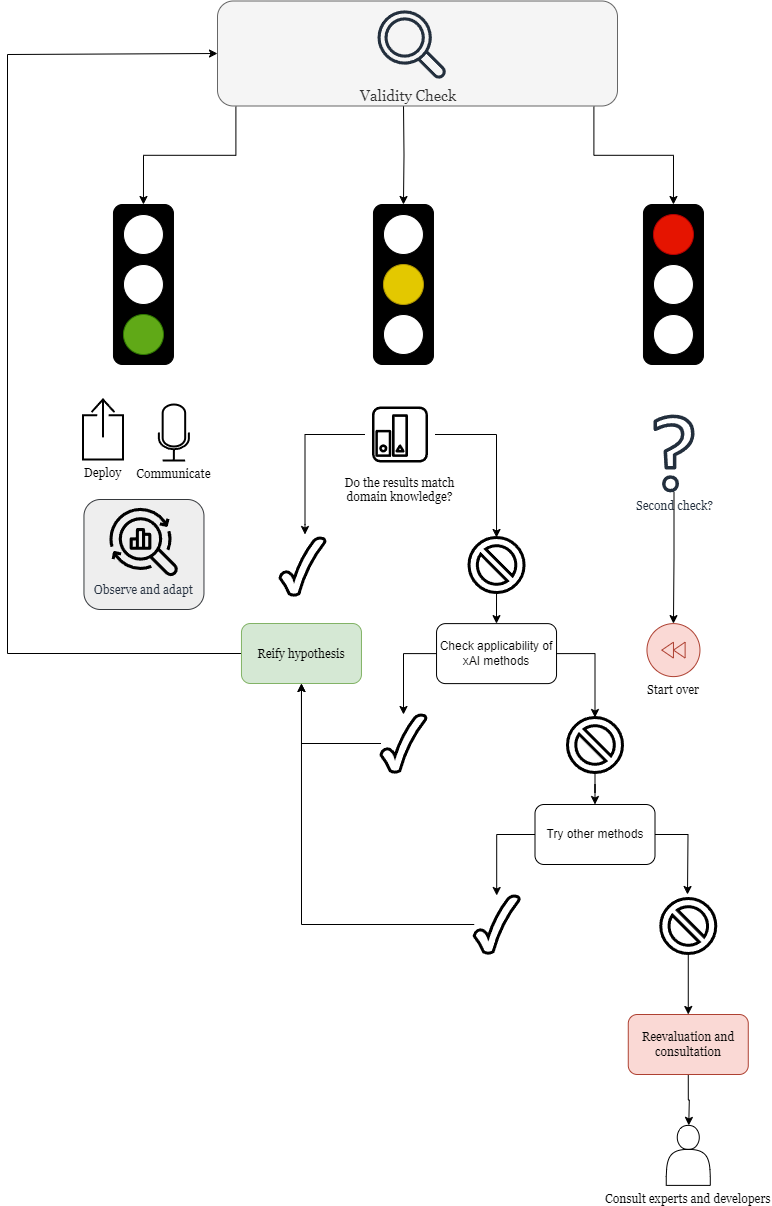}
    \caption{The validity check consists of three scenarios ranging from valid hypothesis to partly valid to not valid.}
    \label{fig:validitycheck}
\end{figure}

If the hypotheses align with xAI results, the explanations can be deployed and communicated (Scenario 1; RQ 2.2). The journey is not over here, though. Continuous observation and adaption are crucial.
If the hypotheses are partly valid, consider the descriptive statistics and domain-knowledge gathered before. If these support the results of the xAI methods, you may need to reify the hypotheses.
If the hypotheses do not align with xAI results and you checked them before as well, it is time to either question the validity of the used xAI method or revise the AI-model itself. Here, you may need to consult domain-experts and developers. 

The proposed framework is generic in the sense that it can be applied in various business scenarios involving xAI. Although, it suffers from several limitations. 

\subsection{Limitations and future research}
As long as AI remains an ever advancing technology, xAI will play an important role. Our proposed framework shows how to approach xRL in a production scheduling setting. 
However, this approach still involves experts manually constructing and comparing the hypotheses to the explanations. 

First of all, the way hypotheses generation is approached needs to be formalized.
There is the need to define objective, transparent criteria for the determinants understandability, comparability, communicability, transferability, and validability of hypotheses in any scenario. The question of the "ideal" hypothesis still persists, because this is a highly complex question. Different factors from various scientific disciplines play a role here. Ideality is a subjective question. 

Second of all, precise rules to identify significant deviations from the hypotheses are lacking, which might be constructed similarly to hypotheses testing in statistics. This would be needed so that ideally the agent may be able to self-explain.

Thirdly, we were not able to use the original custom class NN with the xAI methods and had to rebuild it, thereby introducing some noise. There were 5 cases, where our rebuild network did not match the predictions of the original one. 
While we are inspecting overall patterns, there were individual cases where effects were inconsistent with our hypotheses. Additionally, as discussed in the literature review, there were few unrealistic points in the SHAP plots, most likely due to feature dependence.

In the future, it could be promising to create a self-explaining AI using our hypotheses-based approach. An application that adjusts the type of explanation to the needs of the explainee in a conversational or interactive style might be beneficial to increase understanding. Here, it might be fruitful to employ large language models. Additionally, future research should focus on evaluating the explanations and putting the human – the explainee – in the foreground. Our paper focuses on creating explanations and interpretations, but these should also be tested and adapted in terms of effectiveness, user satisfaction, and trust, which requires studies involving humans and a collaboration with social and cognitive science. \shortcite{kim2021multi,mohseni202124, milani2024explainable}

Also, quantitative metrics to evaluate xRL methods, f.e. regarding their fidelity, are lacking in literature \shortcite{milani2024explainable, xiong2024xrl} and were also not assessed in this paper. In the future, these types of metrics should be included.

The framework does not cover details of approaching stakeholders. There may be several blockades to overcome when communicating xAI. These blockades may be gridlocked opinions in the company, a lack of authority or a lack of productive capital to implement suggested changes (social proof, authority, scarcity; c.f. \shortcite{cialdini2001science}). This should also anticipate human biases and heuristics of thought \shortcite{tversky1974judgment}. Humans prefer explanations which intuitively make sense to them and which align with their world view \shortcite{gentner2014mental}. Workflows should reflect on that to create explanations that are transparent and ready to communicate to third parties \shortcite{riveiro2021s}. 

Our workflow has not yet been tested in other production and manufacturing settings (e.g. the chemical industry), which could be an opportunity for future research. Also, the framework is data-driven and will be hard to implement when available data is scarce.

Lastly, using our hypotheses approach, we consider the causal understanding in the production planning context. However, the xAI methods themselves are primarily of correlative nature and thus cannot provide evidence for causal effects. 
\section{Conclusion}
In this case study, we investigated the application of state-of-the-art xAI techniques for a DRL-based scheduling model. We built upon an existing agent in a real-world flow production setting, focusing on enhancing the interpretability of the agent's decisions for domain experts. Our comprehensive investigation addressed both method-specific questions (RQ1.1, RQ1.2) as well as organizational aspects (RQ2.1, RQ2.2). On the method side, we utilized two prominent xAI frameworks, SHAP (DeepSHAP) and Captum (Input x Gradient), to analyze the reasoning behind the scheduling decisions. On the organizational side, our proposed xAI approach is based on the workflow of \citeA{tchuente2024methodological}, serving as a general procedural model for applying xAI methods in business use cases.

In summary, our findings highlight several critical issues in the current xAI literature, including a lack of falsifiability and consistent terminology, insufficient consideration of domain knowledge, inadequate attention to the target audience or real-world scenarios, and a tendency to offer simple input-output explanations rather than causal interpretations. Moreover, we observed that existing workflows often lack sufficient detail on how xAI methods and their results can be integrated with domain-specific aspects.

To address these challenges, we introduced a hypotheses-based workflow with feedback loops. This workflow allows for the inspection of explanations to ensure they are consistent with domain knowledge and the reward hypotheses of the agent. Our results show that both DeepSHAP and Input x Gradient are well-suited to explain the behavior of the agent, provided the methods are systematically embedded in the proposed workflow. However, DeepSHAP proved to be slightly more effective in our use case, as it was able to differentiate all the agent's actions more clearly. We hypothesize that this xAI workflow may also be applicable to other DRL-based scheduling models and should be tested and further developed in future studies.

\section*{Funding} 
This work was partly supported by the Ministry of Economic Affairs,
Industry, Climate Action and Energy of the State of North Rhine-Westphalia,
Germany, under the project SUPPORT (005-2111-0026)

\section*{Author contributions (CRediT)}

Daniel Fischer: \textit{Methodology, Formal analysis, Investigation, Software, Writing - original draft};
Hannah M. Hüsener: \textit{Methodology, Formal analysis, Investigation, Software, Writing - original draft}; 
Felix Grumbach: \textit{Conceptualization, Supervision, Writing – review \& editing}; 
Lukas Vollenkemper: \textit{Conceptualization, Supervision, Writing – review \& editing}; 
Arthur Müller: \textit{Data curation, Software, Writing – review \& editing};
Pascal Reusch: \textit{Funding acquisition, Resources}; 

\section*{Declarations}
\subsection*{Conflict of interest}
The authors have no conflicts of interest to declare that are relevant to the content of this article.

\subsection*{Open Access}
This article is licensed under a Creative Commons Attribution 4.0 International License, which permits use, sharing, adaptation, distribution and reproduction in any medium or format, as long as you give appropriate credit to the original author(s) and the source, provide a link to the Creative Commons licence, and indicate if changes were made. The images or other third party material in this article are included in the article’s Creative Commons licence, unless indicated otherwise in a credit line to the material. If material is not included in the article’s Creative Commons licence and your intended use is not permitted by statutory regulation or exceeds the permitted use, you will need to obtain permission directly from the copyright holder. To view a copy of this licence, visit http://creativecommons.org/licenses/by/4.0/.
\bibliographystyle{apacite}
\bibliography{ref} 
\clearpage
\markboth{Appendix}{Appendix}
\section{Appendix}

\label{Appendix_InputxGradient}
\subsection{Input x Gradient Attribution Plots}

\begin{figure}[H]
    \centering
    \includegraphics[width=1\linewidth]{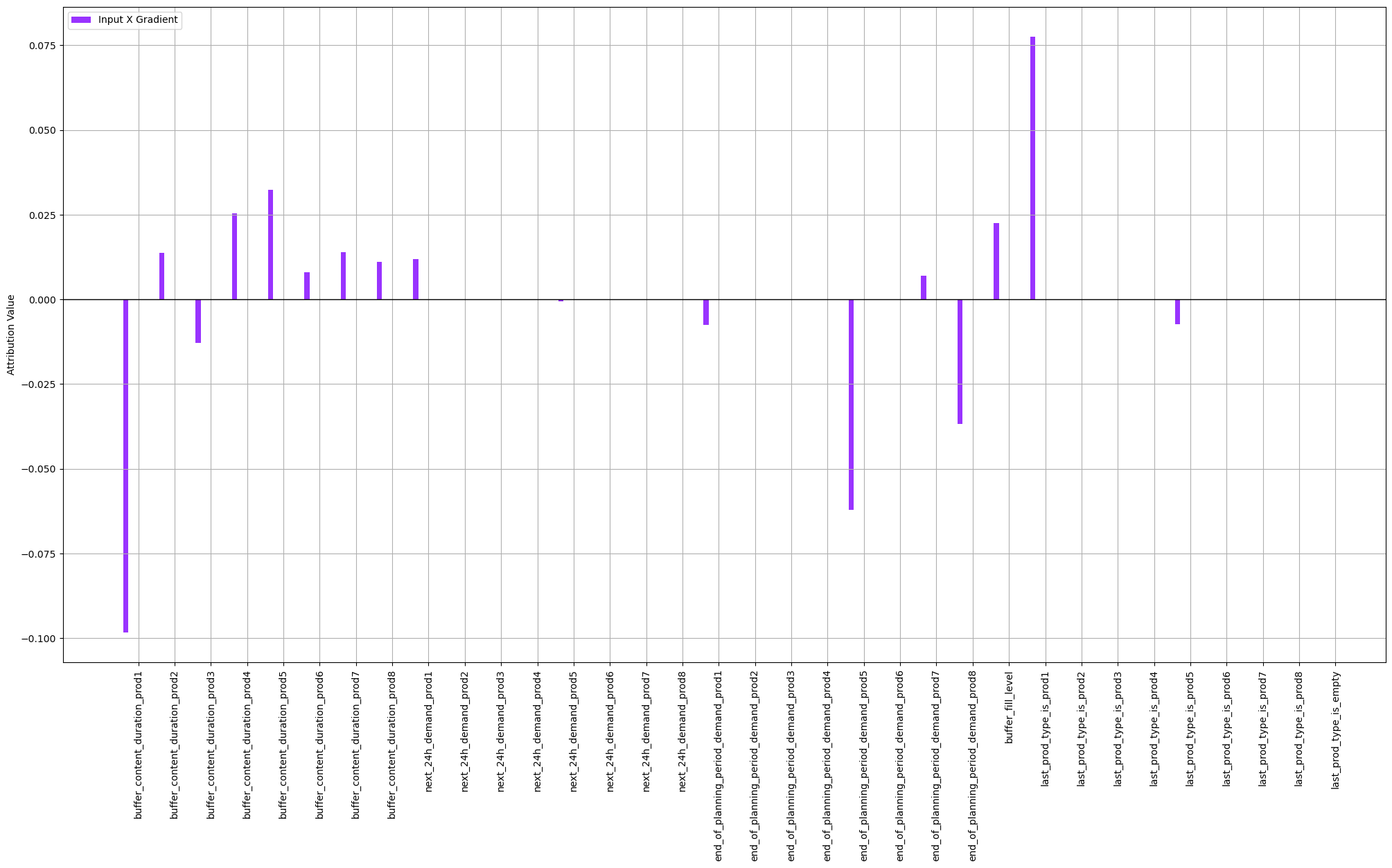}
    \caption{Attribution plot for Product 1}
    \label{fig:appenidix_agg1_Captum}
\end{figure}

\begin{figure}[H]
    \centering
    \includegraphics[width=1\linewidth]{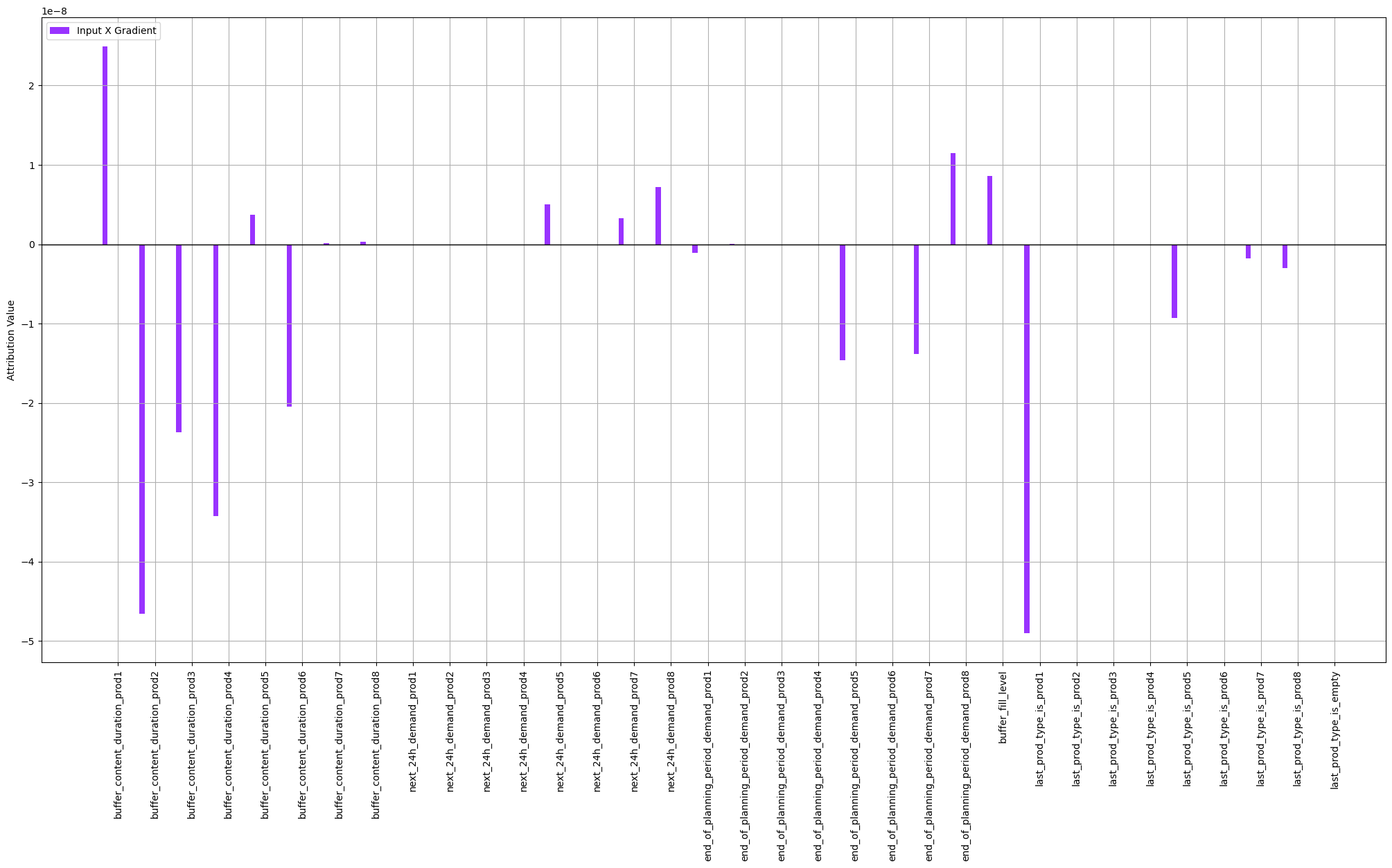}
    \caption{Attribution plot for Product 2}
    \label{fig:appenidix_agg2_Captum}
\end{figure}

\begin{figure}[H]
    \centering
    \includegraphics[width=1\linewidth]{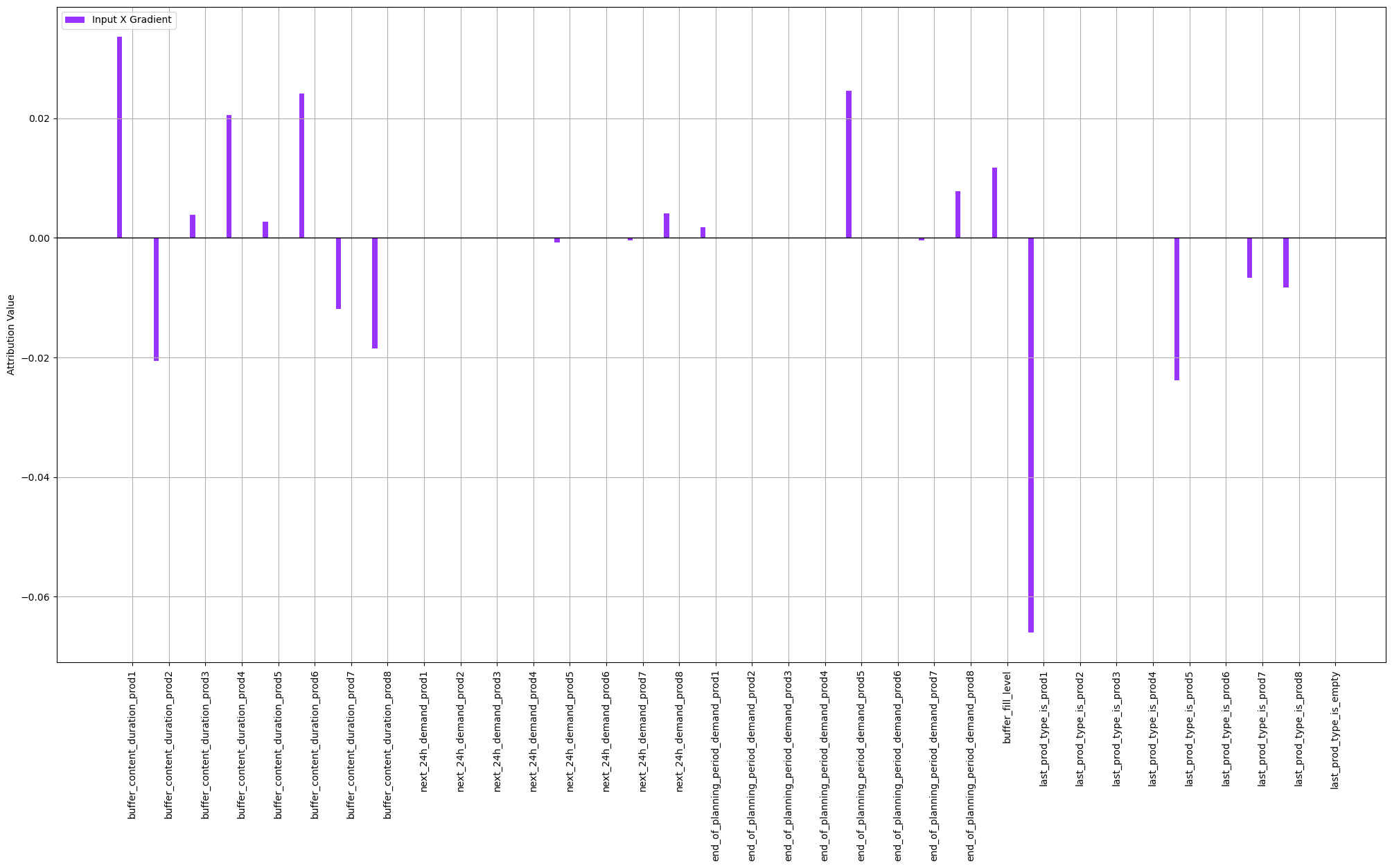}
    \caption{Attribution plot for Product 3}
    \label{fig:appenidix_agg3_Captum}
\end{figure}

\begin{figure}[H]
    \centering
    \includegraphics[width=1\linewidth]{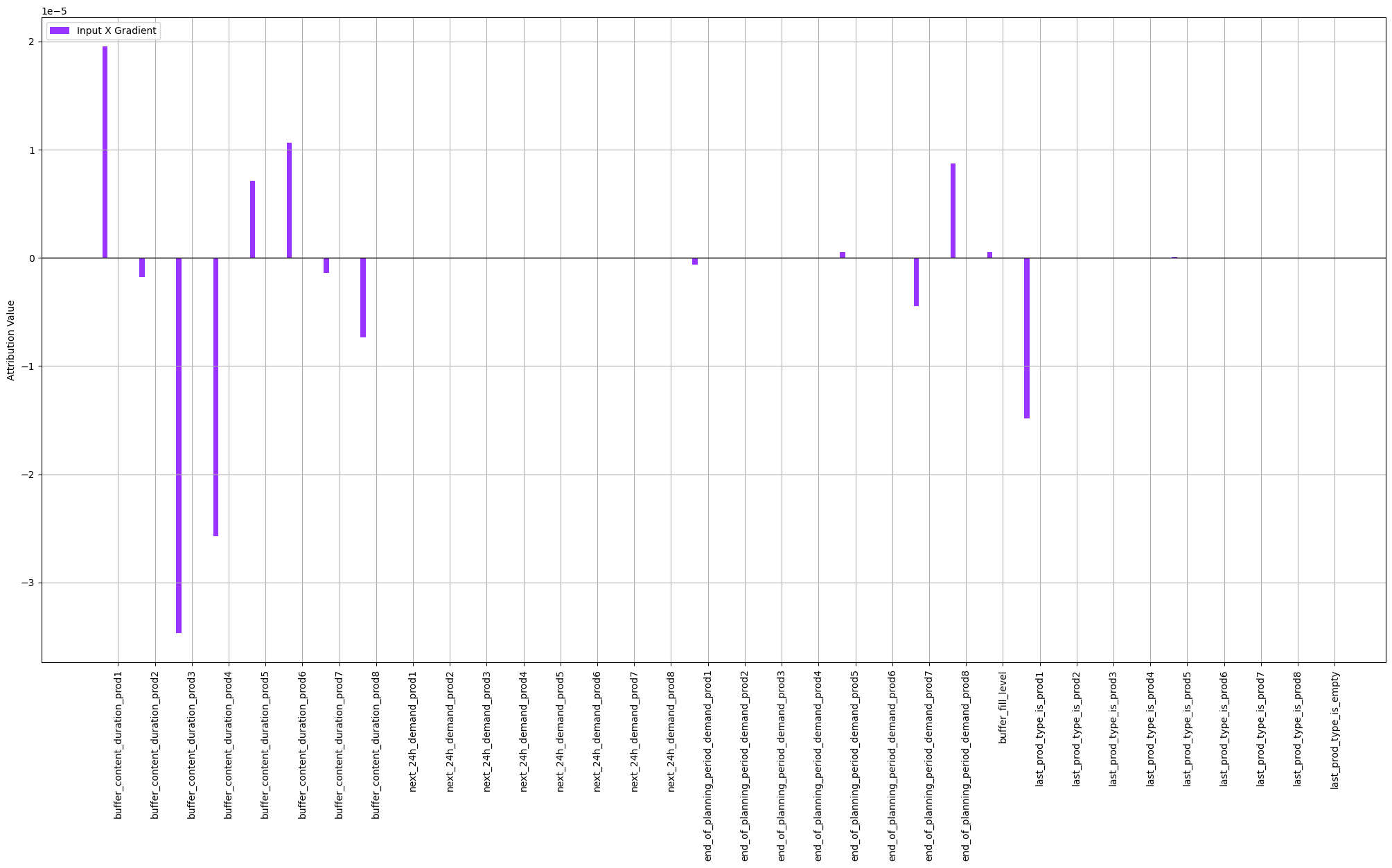}
    \caption{Attribution plot for Product 4}
    \label{fig:appenidix_agg4_Captum}
\end{figure}

\begin{figure}[H]
    \centering
    \includegraphics[width=1\linewidth]{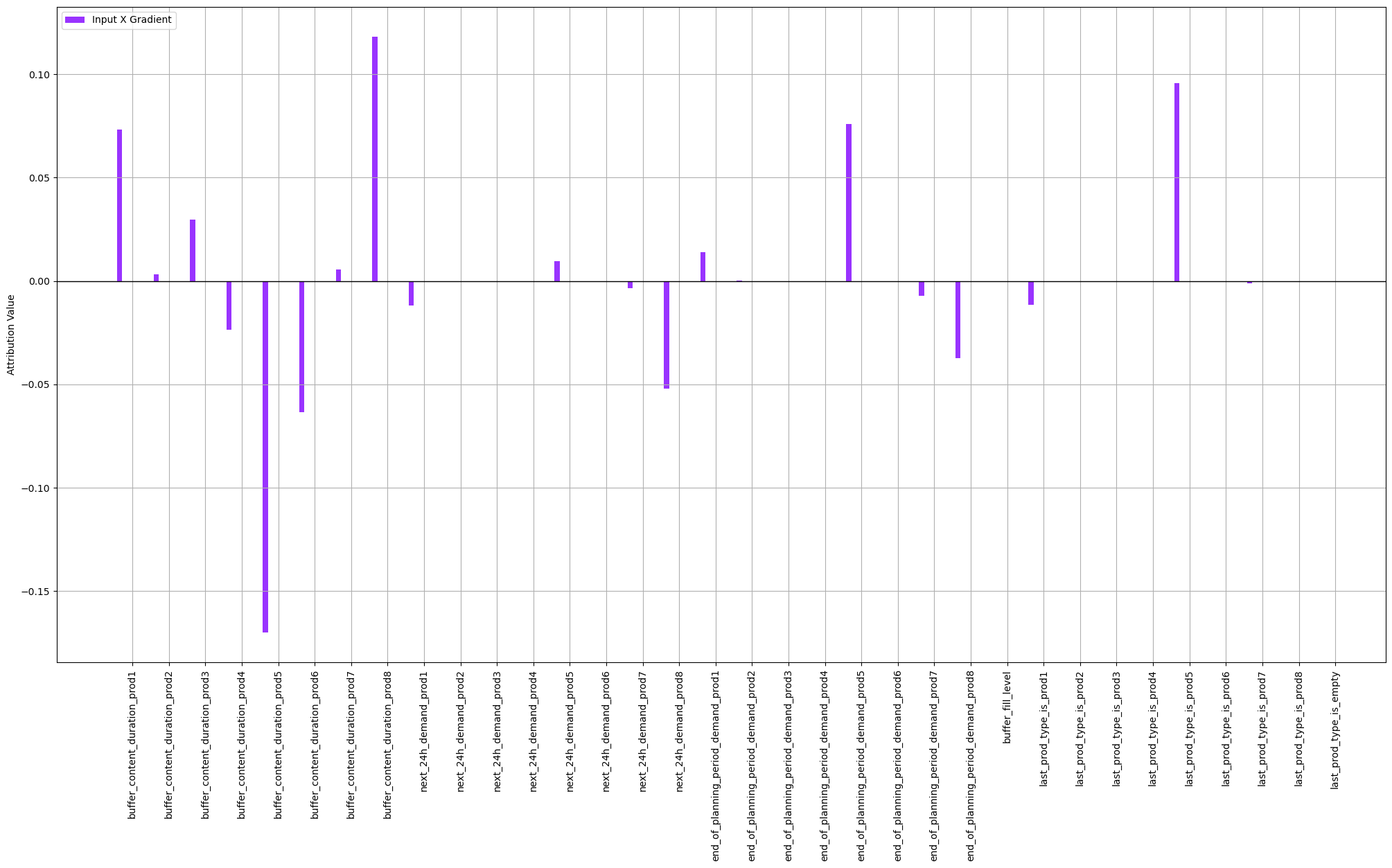}
    \caption{Attribution plot for Product 5}
    \label{fig:appenidix_agg5_Captum}
\end{figure}

\begin{figure}[H]
    \centering
    \includegraphics[width=1\linewidth]{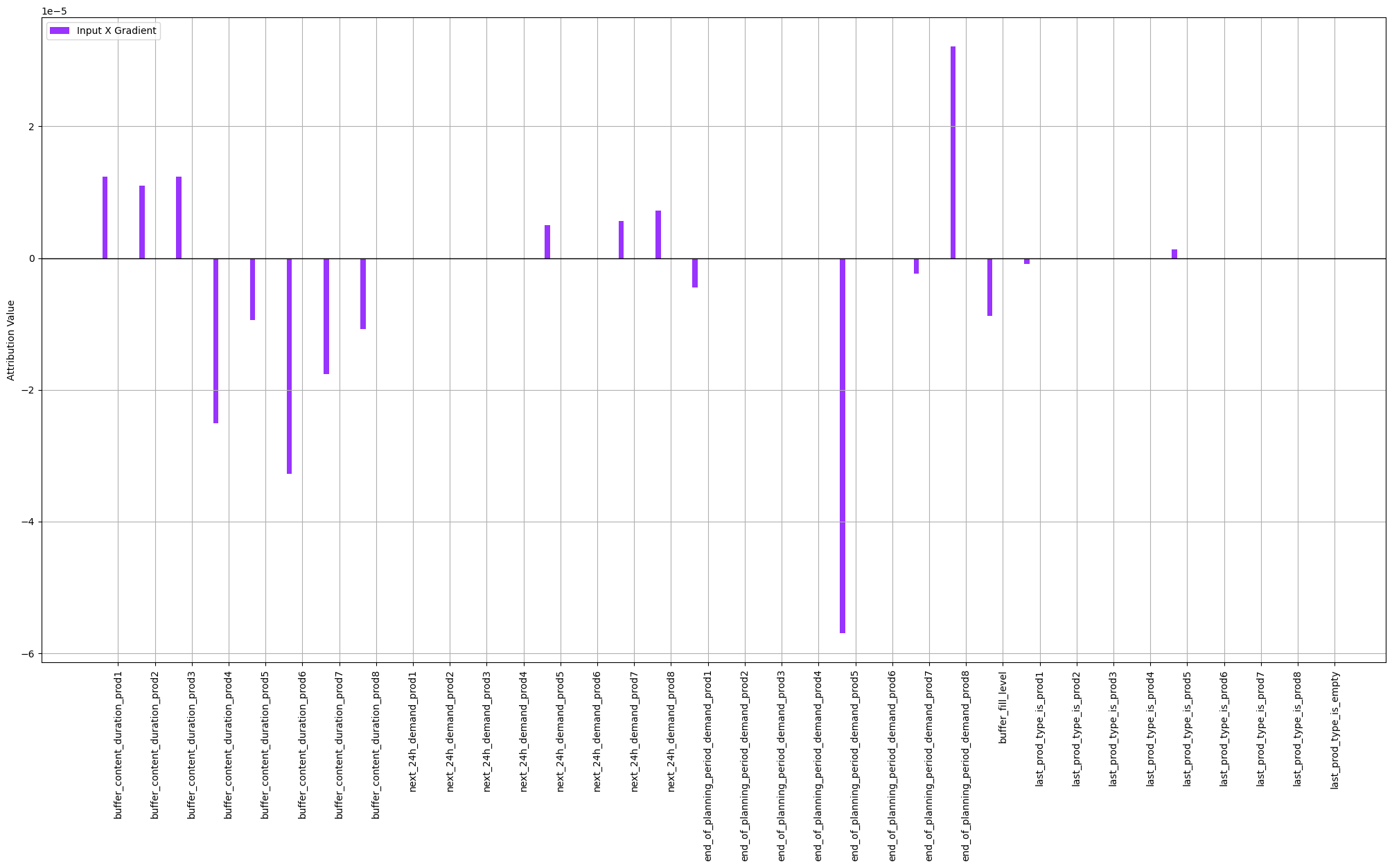}
    \caption{Attribution plot for Product 6}
    \label{fig:appenidix_agg6_Captum}
\end{figure}

\begin{figure}[H]
    \centering
    \includegraphics[width=1\linewidth]{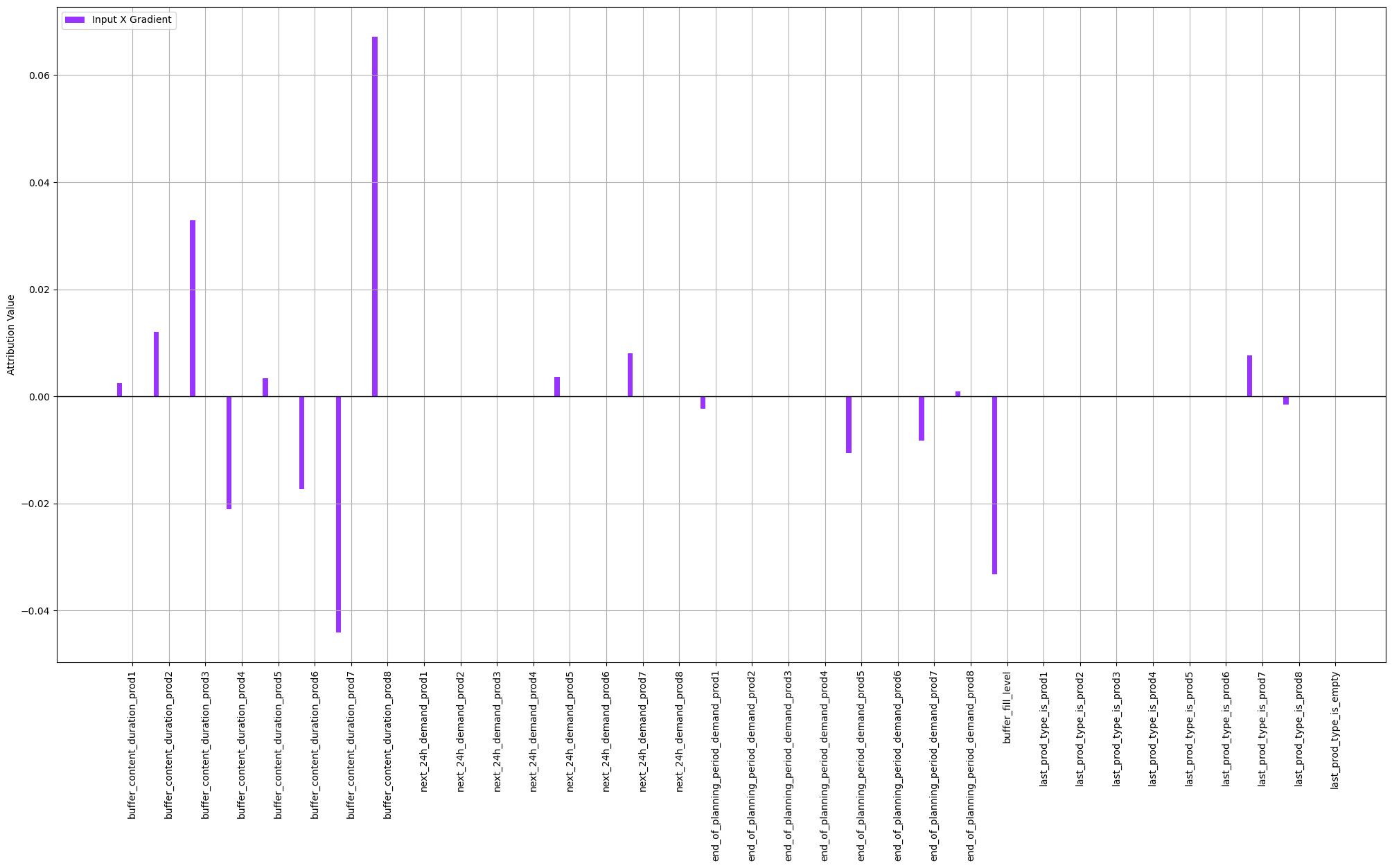}
    \caption{Attribution plot for Product 7}
    \label{fig:appenidix_agg7_Captum}
\end{figure}

\begin{figure}[H]
    \centering
    \includegraphics[width=1\linewidth]{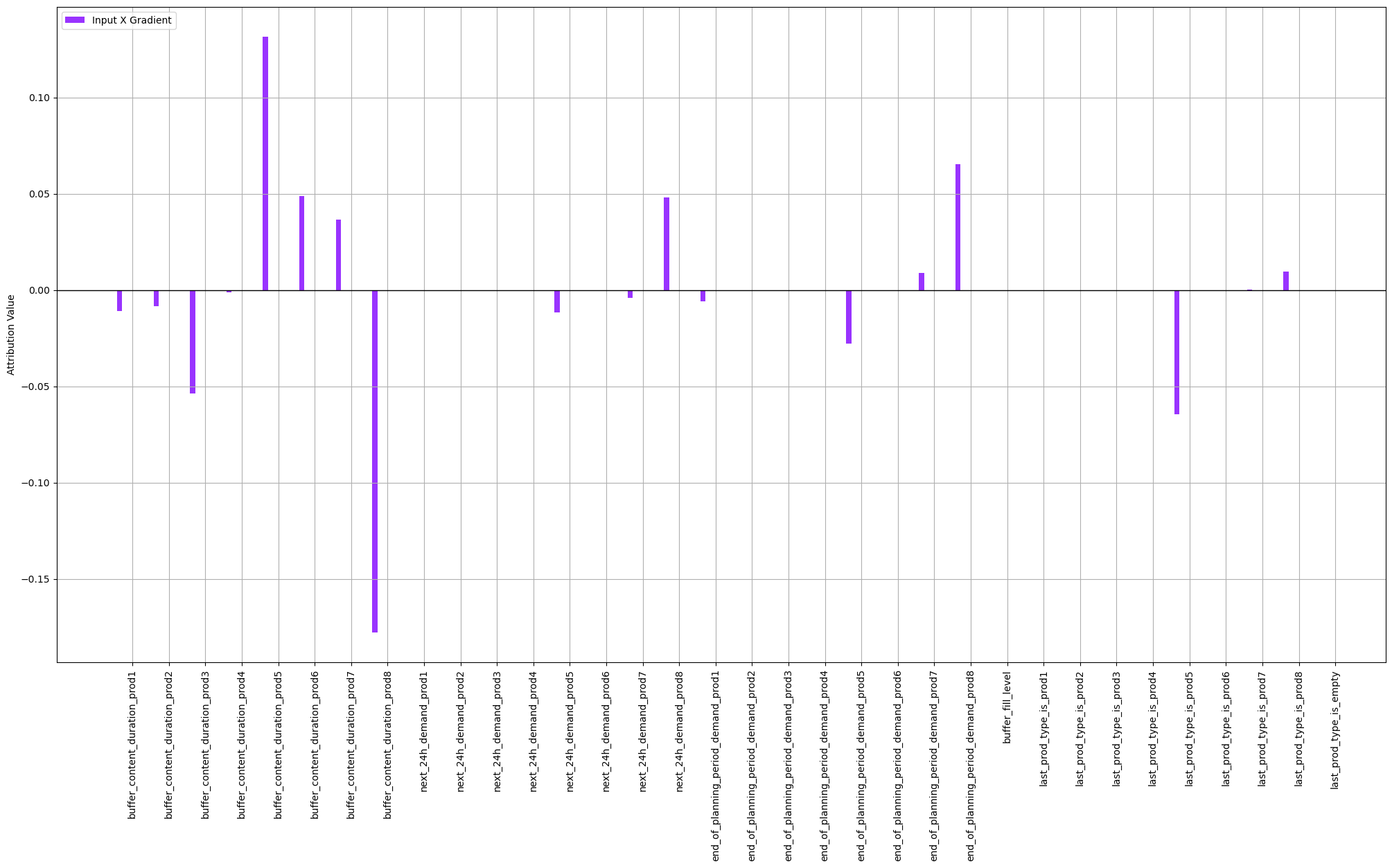}
    \caption{Attribution plot for Product 8}
    \label{fig:appenidix_agg8_Captum}
\end{figure}

\label{Appendix_SHAP}
\subsection{Product 1}

\begin{table}[ht!]
    \footnotesize
    \centering
    \caption{Top 10 variables for Input X Gradient and Product 1.}
    \label{tab:variables}
    \begin{tabularx}{\textwidth}{lXr}
        \toprule
        \textbf{Ranking} & \textbf{Variable} & \textbf{Value} \\
        \midrule
        1  & buffer\_content\_duration\_prod1 & -0.09835365414619446 \\
        2  & last\_prod\_type\_is\_prod1 & 0.07755021005868912 \\
        3  & end\_of\_planning\_period\_demand\_prod5 & -0.06204185262322426 \\
        4  & end\_of\_planning\_period\_demand\_prod8 & -0.03671699017286301 \\
        5  & buffer\_content\_duration\_prod5 & 0.032430436462163925 \\
        6  & buffer\_content\_duration\_prod4 & 0.025369716808199883 \\
        7  & buffer\_fill\_level & 0.022456379607319832 \\
        8  & buffer\_content\_duration\_prod7 & 0.013850999064743519 \\
        9  & buffer\_content\_duration\_prod2 & 0.01368733774870634 \\
        10 & buffer\_content\_duration\_prod3 & -0.012776759453117847 \\
        \bottomrule
    \end{tabularx}
\end{table}
\FloatBarrier

\label{Appendix_Notproduced_Aggs}
\subsection{Product 6}

Product 6 was never produced.

\begin{figure}[H]
    \centering
    \includegraphics[width=0.75\linewidth]{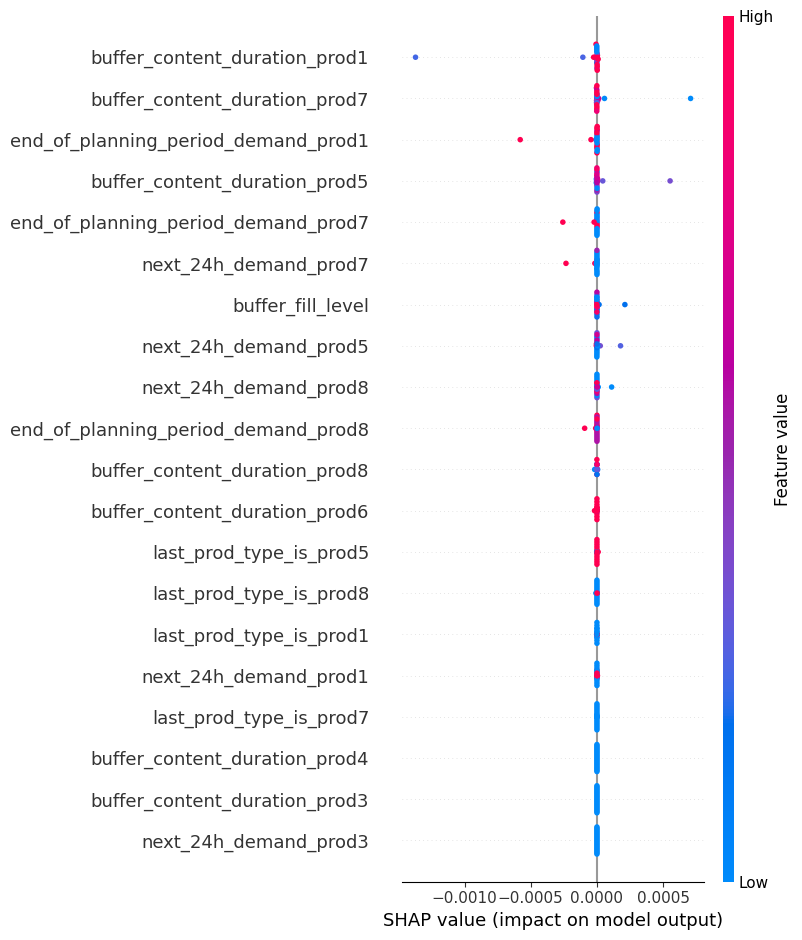}
    \caption{SHAP summary plot for action 5. 
Each point represents the local feature attribution value (Shapley value for feature and instance).
Blue color indicates a low feature value, for binary variables this is 0, red indicates high feature values, for binary variables this is 1. A positive SHAP value is positively associated with the action, a negative SHAP value is negatively associated with the action. The features are displayed based on importance on average with decreasing importance from top to bottom.}
    \label{fig:SHAP_Action5}
\end{figure}

Regarding SHAP, we can see that all variables have SHAP values around zero (\ref{fig:SHAP_Action5}). 

These are the attributions for Input x Gradient. As again, the values are too small to interpret them:

\begin{table}[ht!]
    \footnotesize
    \centering
    \caption{Top 10 Variables for Input X Gradient and Product 6}
    \label{tab:top_variables_target5}
    \begin{tabularx}{\textwidth}{lXr}
        \toprule
        \textbf{Rank} & \textbf{Variable} & \textbf{Value} \\
        \midrule
        1  & end\_of\_planning\_period\_demand\_prod5 & -5.6928e-05 \\
        2  & buffer\_content\_duration\_prod6 & -3.2762e-05 \\
        3  & end\_of\_planning\_period\_demand\_prod8 & 3.2095e-05 \\
        4  & buffer\_content\_duration\_prod4 & -2.5057e-05 \\
        5  & buffer\_content\_duration\_prod7 & -1.7585e-05 \\
        6  & buffer\_content\_duration\_prod3 & 1.2410e-05 \\
        7  & buffer\_content\_duration\_prod1 & 1.2367e-05 \\
        8  & buffer\_content\_duration\_prod2 & 1.0998e-05 \\
        9  & buffer\_content\_duration\_prod8 & -1.0736e-05 \\
        10 & buffer\_content\_duration\_prod5 & -9.3647e-06 \\
        \bottomrule
    \end{tabularx}
\end{table}
\FloatBarrier

The attribution values are too low to be interpreted.

In summary, for both methods we can support the hypothesis that products that were not produced should play a less important role in xAI methods. 

\subsection{Product 2}

Product 2 was never produced.  

\begin{figure}[H]
    \centering
    \includegraphics[width=0.75\linewidth]{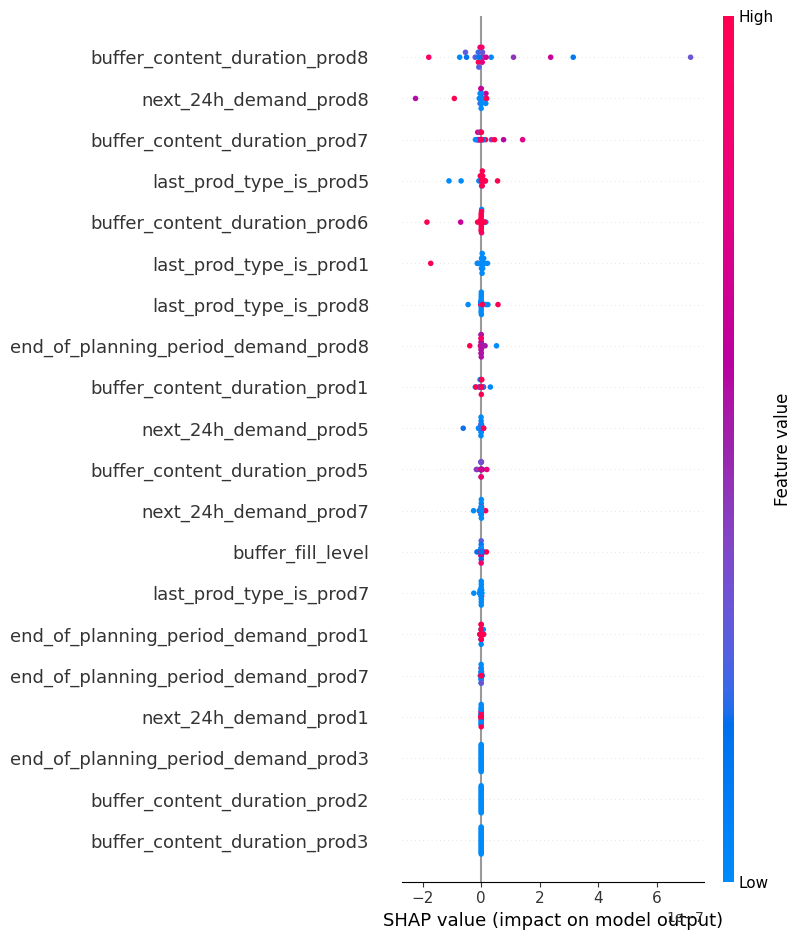}
    \caption{SHAP summary plot for action 1. 
Each point represents the local feature attribution value (Shapley value for feature and instance).
Blue color indicates a low feature value, for binary variables this is 0, red indicates high feature values, for binary variables this is 1. A positive SHAP value is positively associated with the action, a negative SHAP value is negatively associated with the action. The features are displayed based on importance on average with decreasing importance from top to bottom.}
    \label{fig:SHAP_Action1}
\end{figure}

Regarding SHAP, SHAP values for most features are close to zero (\ref{fig:SHAP_Action1}). The trend for the most important variable, buffer content duration of prod8 is ambiguous: Higher buffer content of prod8 is plotted with negative, null, and positive SHAP values simultaneously, which cannot be interpreted.

However, there are few individual points indicating that a higher next 24h demand of prod8 is negatively associated with production of prod1, which might be one of the reasons why this prod was not produced. According to our hypothesis, this would mean that production of prod8 was critical; therefore, it was produced instead of prod2. 

A fuller buffer content of prod7 is slightly associated with production of prod2. Even though, prod2 was not produced, this is still in line with our hypothesis. If prod7 is not critical (e.g., the buffer is full) that makes it more likely that another product, for example prod2 can be produced. However, the SHAP values are close to zero, which makes sense, because the product was not produced. 

Last product type is prod5 is fourth important in this plot. We can see that if the last product produced was not prod5, it speaks against production of prod2. Prod5 was produced the most often; therefore, it makes sense that if prod2 had been produced, the chances would be higher for this to happen following prod5. 

We can also see that buffer content of prod6 and production of prod1 is slightly negatively associated with production of prod2. The other SHAP values for the rest of features are close to zero.  

In line with our hypothesis that the produced products should play are more important role in the xAI methods than the ones that were not produced, using visual inspection we can tell that asides from few outliers most features have SHAP values around zero and - besides buffer content duration for prod8 with an ambiguous pattern - no feature is strongly associated with production of prod2. 
We can see that - again asides from the ambiguous pattern in buffer content duration for prod8 - the few features that show greater absolute SHAP values speak against production of prod2, which makes intuitive sense, because the product was not produced.

The following table shows the Input X Gradient attribution values:

\begin{table}[ht!]
    \footnotesize
    \centering
    \caption{Top 10 Variables for Input X Gradient and Product 2}
    \label{tab:top_variables_target1}
    \begin{tabularx}{\textwidth}{lXr}
        \toprule
        \textbf{Rank} & \textbf{Variable} & \textbf{Value} \\
        \midrule
        1  & last\_prod\_type\_is\_prod1 & -4.8995e-08 \\
        2  & buffer\_content\_duration\_prod2 & -4.6587e-08 \\
        3  & buffer\_content\_duration\_prod4 & -3.4221e-08 \\
        4  & buffer\_content\_duration\_prod1 & 2.4892e-08 \\
        5  & buffer\_content\_duration\_prod3 & -2.3662e-08 \\
        6  & buffer\_content\_duration\_prod6 & -2.0466e-08 \\
        7  & end\_of\_planning\_period\_demand\_prod5 & -1.4639e-08 \\
        8  & end\_of\_planning\_period\_demand\_prod7 & -1.3842e-08 \\
        9  & end\_of\_planning\_period\_demand\_prod8 & 1.1487e-08 \\
        10 & last\_prod\_type\_is\_prod5 & -9.3096e-09 \\
        \bottomrule
    \end{tabularx}
\end{table}
\FloatBarrier

The attribution values are too small to interpret. 

In summary, for both methods the hypothesis that produced products should play a more important role can be supported.

\subsection{Product 3}

Product 3 was also never produced. 

\begin{figure}[H]
    \centering
    \includegraphics[width=0.75\linewidth]{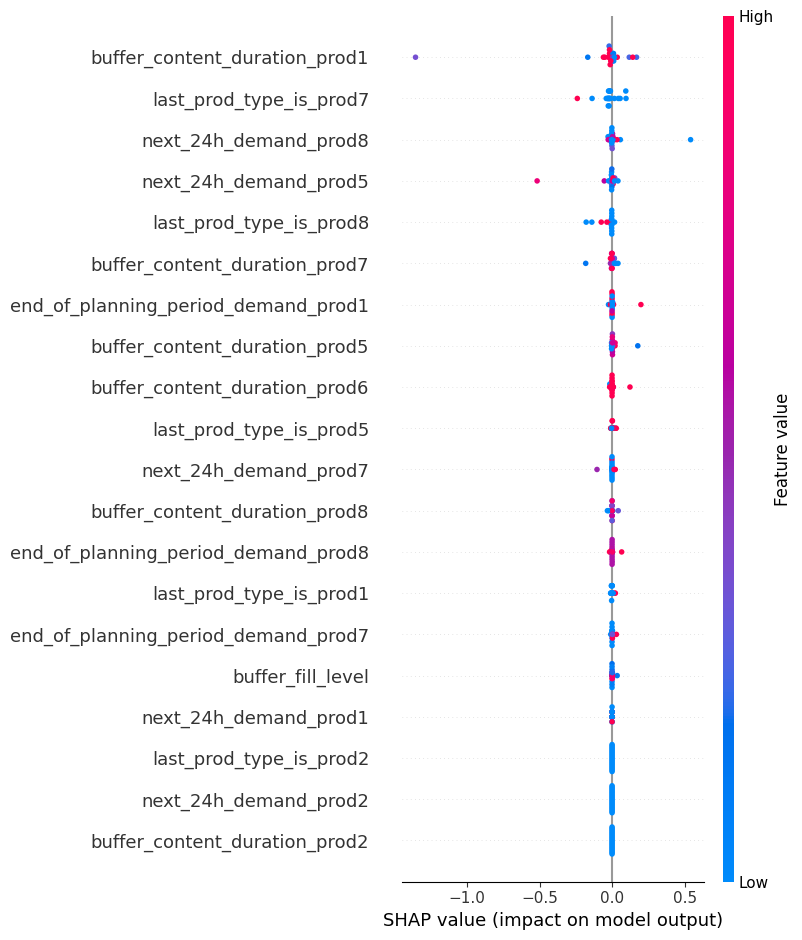}
    \caption{SHAP summary plot for action 2. 
Each point represents the local feature attribution value (Shapley value for feature and instance).
Blue color indicates a low feature value, for binary variables this is 0, red indicates high feature values, for binary variables this is 1. A positive SHAP value is positively associated with the action, a negative SHAP value is negatively associated with the action. The features are displayed based on importance on average with decreasing importance from top to bottom.}
    \label{fig:SHAP_Action2}
\end{figure}

Regarding SHAP, we can see that asides from few individual points most variables have SHAP values around zero and do not strongly speak for production of prod3 (\ref{fig:SHAP_Action2}). 

Regarding Input x Gradient, we derived the following attributions: 

\begin{table}[ht!]
    \footnotesize
    \centering
    \caption{Top 10 Variables for Input X Gradient and Product 3}
    \label{tab:top_variables_target2}
    \begin{tabularx}{\textwidth}{lXr}
        \toprule
        \textbf{Rank} & \textbf{Variable} & \textbf{Value} \\
        \midrule
        1  & last\_prod\_type\_is\_prod1 & -0.06599830090999603 \\
        2  & buffer\_content\_duration\_prod1 & 0.03362700715661049 \\
        3  & end\_of\_planning\_period\_demand\_prod5 & 0.02455740049481392 \\
        4  & buffer\_content\_duration\_prod6 & 0.02413715235888958 \\
        5  & last\_prod\_type\_is\_prod5 & -0.023769624531269073 \\
        6  & buffer\_content\_duration\_prod2 & -0.0206350926309824 \\
        7  & buffer\_content\_duration\_prod4 & 0.020487982779741287 \\
        8  & buffer\_content\_duration\_prod8 & -0.018504632636904716 \\
        9  & buffer\_content\_duration\_prod7 & -0.01190112717449665 \\
        10 & buffer\_fill\_level & 0.011671244166791439 \\
        \bottomrule
    \end{tabularx}
\end{table}
\FloatBarrier

If the product before was product 1, the chances that product 3 will be produced again are decreased.

In summary, for SHAP we can support that hypothesis that products that were not produced should play a less important role in xAI methods. 

\subsection{Product 4}

Product 4 was never produced.

\begin{figure}[H]
    \centering
    \includegraphics[width=0.75\linewidth]{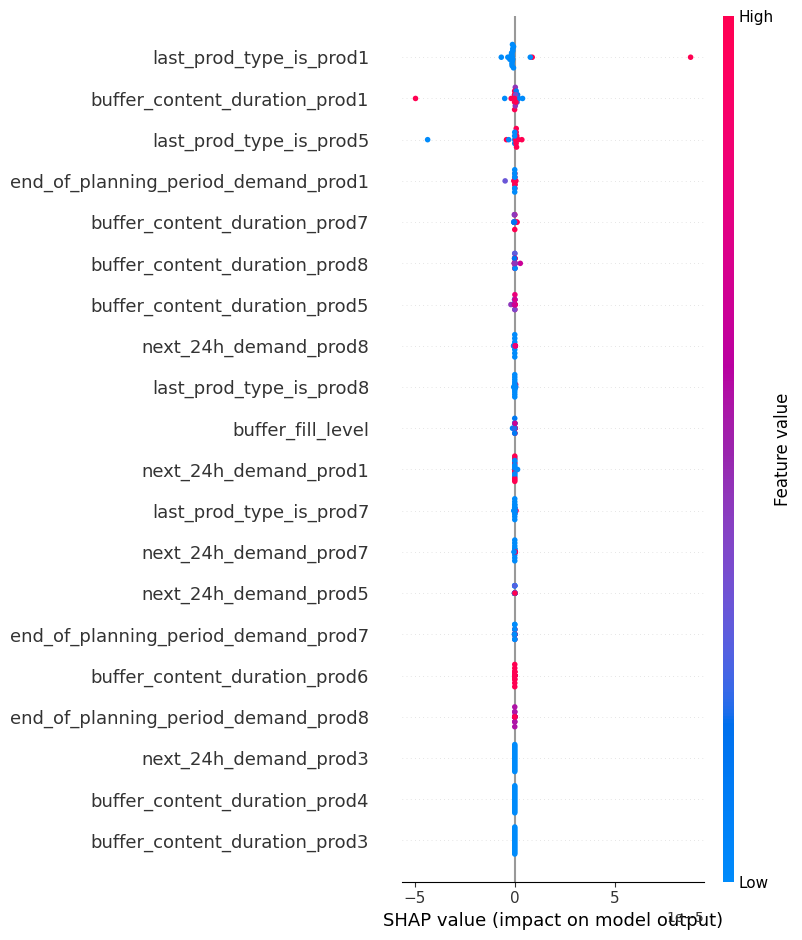}
    \caption{SHAP summary plot for action 3. 
Each point represents the local feature attribution value (Shapley value for feature and instance).
Blue color indicates a low feature value, for binary variables this is 0, red indicates high feature values, for binary variables this is 1. A positive SHAP value is positively associated with the action, a negative SHAP value is negatively associated with the action. The features are displayed based on importance on average with decreasing importance from top to bottom.}
    \label{fig:SHAP_Action3}
\end{figure}

Regarding SHAP, we can see that asides from few individual points most variables have SHAP values around zero and do not strongly speak for production of prod4 (\ref{fig:SHAP_Action3}). 

Regarding Input x Gradient, we derive

\begin{table}[ht!]
    \footnotesize
    \centering
    \caption{Top 10 Variables for Input X Gradient and Product 4}
    \label{tab:top_variables_target3}
    \begin{tabularx}{\textwidth}{lXr}
        \toprule
        \textbf{Rank} & \textbf{Variable} & \textbf{Value} \\
        \midrule
        1  & buffer\_content\_duration\_prod3 & -3.4675e-05 \\
        2  & buffer\_content\_duration\_prod4 & -2.5701e-05 \\
        3  & buffer\_content\_duration\_prod1 & 1.9520e-05 \\
        4  & last\_prod\_type\_is\_prod1 & -1.4845e-05 \\
        5  & buffer\_content\_duration\_prod6 & 1.0646e-05 \\
        6  & end\_of\_planning\_period\_demand\_prod8 & 8.7112e-06 \\
        7  & buffer\_content\_duration\_prod8 & -7.3491e-06 \\
        8  & buffer\_content\_duration\_prod5 & 7.1326e-06 \\
        9  & end\_of\_planning\_period\_demand\_prod7 & -4.4811e-06 \\
        10 & buffer\_content\_duration\_prod2 & -1.7606e-06 \\
        \bottomrule
    \end{tabularx}
\end{table}
\FloatBarrier

The attribution values are too small to interpret.

In summary, for both methods we can support the that hypothesis products that were not produced should play a less important role in xAI methods. 

\subsection{Product 7}

Regarding Input x Gradient, these are the top ten attributions: 

\begin{table}[ht!]
    \footnotesize
    \centering
    \caption{Top 10 Variables for Input X Gradient and Product 7}
    \label{tab:top_variables_target6}
    \begin{tabularx}{\textwidth}{lXr}
        \toprule
        \textbf{Rank} & \textbf{Variable} & \textbf{Value} \\
        \midrule
        1  & buffer\_content\_duration\_prod8 & 0.06717950850725174 \\
        2  & buffer\_content\_duration\_prod7 & -0.0441112294793129 \\
        3  & buffer\_fill\_level & -0.033247679471969604 \\
        4  & buffer\_content\_duration\_prod3 & 0.032851435244083405 \\
        5  & buffer\_content\_duration\_prod4 & -0.02102786675095558 \\
        6  & buffer\_content\_duration\_prod6 & -0.01731930486857891 \\
        7  & buffer\_content\_duration\_prod2 & 0.012098127976059914 \\
        8  & end\_of\_planning\_period\_demand\_prod5 & -0.010609464719891548 \\
        9  & end\_of\_planning\_period\_demand\_prod7 & -0.008218199014663696 \\
        10 & next\_24h\_demand\_prod7 & 0.008046381175518036 \\
        \bottomrule
    \end{tabularx}
\end{table}
\FloatBarrier

\subsection{Product 8}

Regarding Input x Gradient, these were the attributions:

\begin{table}[ht!]
    \footnotesize
    \centering
    \caption{Top 10 Variables for Input X Gradient and Product 8}
    \label{tab:top_variables_target7}
    \begin{tabularx}{\textwidth}{lXr}
        \toprule
        \textbf{Rank} & \textbf{Variable} & \textbf{Value} \\
        \midrule
        1  & buffer\_content\_duration\_prod8 & -0.17791211605072021 \\
        2  & buffer\_content\_duration\_prod5 & 0.13158348202705383 \\
        3  & end\_of\_planning\_period\_demand\_prod8 & 0.0651685819029808 \\
        4  & last\_prod\_type\_is\_prod5 & -0.06460312008857727 \\
        5  & buffer\_content\_duration\_prod3 & -0.0536046028137207 \\
        6  & buffer\_content\_duration\_prod6 & 0.04870598390698433 \\
        7  & next\_24h\_demand\_prod8 & 0.04792417585849762 \\
        8  & buffer\_content\_duration\_prod7 & 0.0367254838347435 \\
        9  & end\_of\_planning\_period\_demand\_prod5 & -0.027777016162872314 \\
        10 & next\_24h\_demand\_prod5 & -0.011823729611933231 \\
        \bottomrule
    \end{tabularx}
\end{table}
\FloatBarrier

\subsection{Robustness check}
Using week 42 from real-world production, we generated new plots using Input x Gradient and SHAP for this new data. Then, we analyzed if our hypotheses still hold true, in order to ensure robustness of our approach.

In week 42, product 4 batches were produced 37 times and product 5 batches were produced 30 times. There were only two cases where our rebuild-network did not match the original one.
\begin{figure}
    \centering
    \includegraphics[width=1\linewidth]{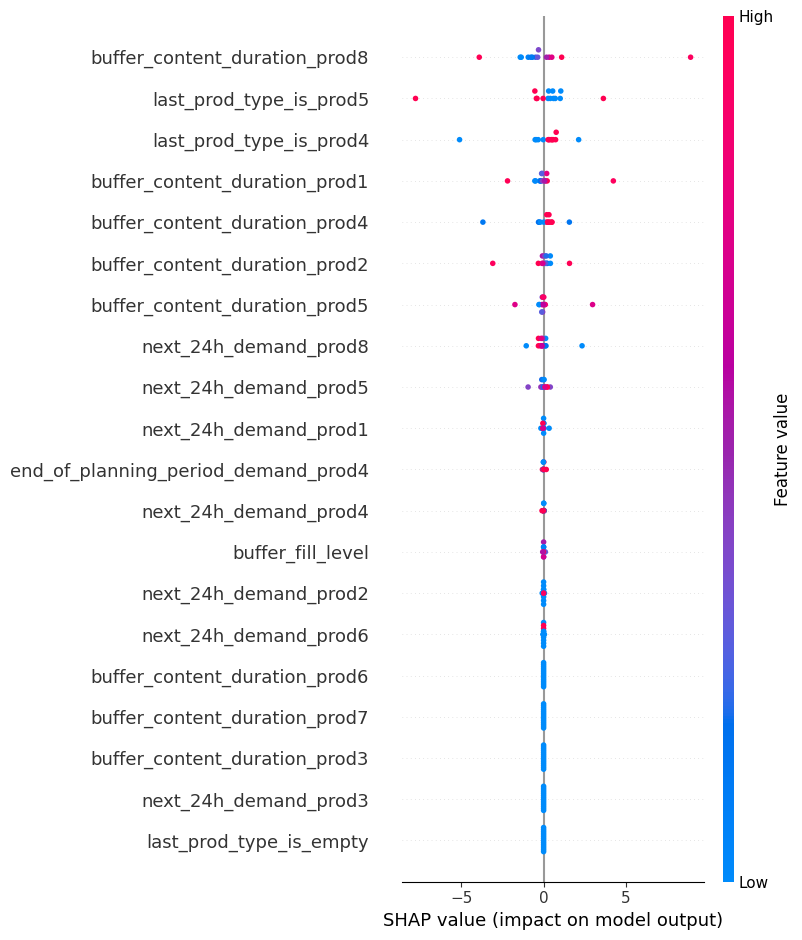}
    \caption{SHAP plot for product 4.}
    \label{fig:agg4-week42}
\end{figure}

\begin{table}[ht!]
    \footnotesize
    \centering
    \caption{Top 10 Variables for Input X Gradient prod4}
    \label{tab:top_variables_prod4}
    \begin{tabularx}{\textwidth}{lXr}
        \toprule
        \textbf{Rank} & \textbf{Variable} & \textbf{Value} \\
        \midrule
        1  & end\_of\_planning\_period\_demand\_prod8 & -4.196784253451824e-09 \\
        2  & buffer\_content\_duration\_prod1 & -2.7213378217538775e-09 \\
        3  & buffer\_fill\_level & 2.443531599283233e-09 \\
        4  & buffer\_content\_duration\_prod3 & -1.8710877291994166e-09 \\
        5  & end\_of\_planning\_period\_demand\_prod5 & -1.3332452919456728e-09 \\
        6  & buffer\_content\_duration\_prod4 & -9.293499303453245e-10 \\
        7  & buffer\_content\_duration\_prod5 & 8.536242268597505e-10 \\
        8  & buffer\_content\_duration\_prod7 & 7.789259237611645e-10 \\
        9  & end\_of\_planning\_period\_demand\_prod7 & -3.3234726082298494e-10 \\
        10 & next\_24h\_demand\_prod5 & -2.8125202167217367e-10 \\
        \bottomrule
    \end{tabularx}
\end{table}
\FloatBarrier

Except one outlier, more buffer content of product 8 makes production of product 4 more likely (criticality-hypothesis).
Except one outlier, if product 4 was produced last, it is more likely to be produced again (setup-effort-hypothesis).
For buffer content duration of product 4 the trend is ambiguous, but a fuller buffer seems to speak for production. While this goes against our criticality hypothesis, it is inline with minimizing setup efforts; product 4 was produced 37 times, 36 of these without interruption. If no other product was critical (most variables for demand are close to zero), it makes sense that the agent stuck to production of product 4, even though the buffer content was already fuller, in order to minimize setup times.
\begin{figure}
    \centering
    \includegraphics[width=1\linewidth]{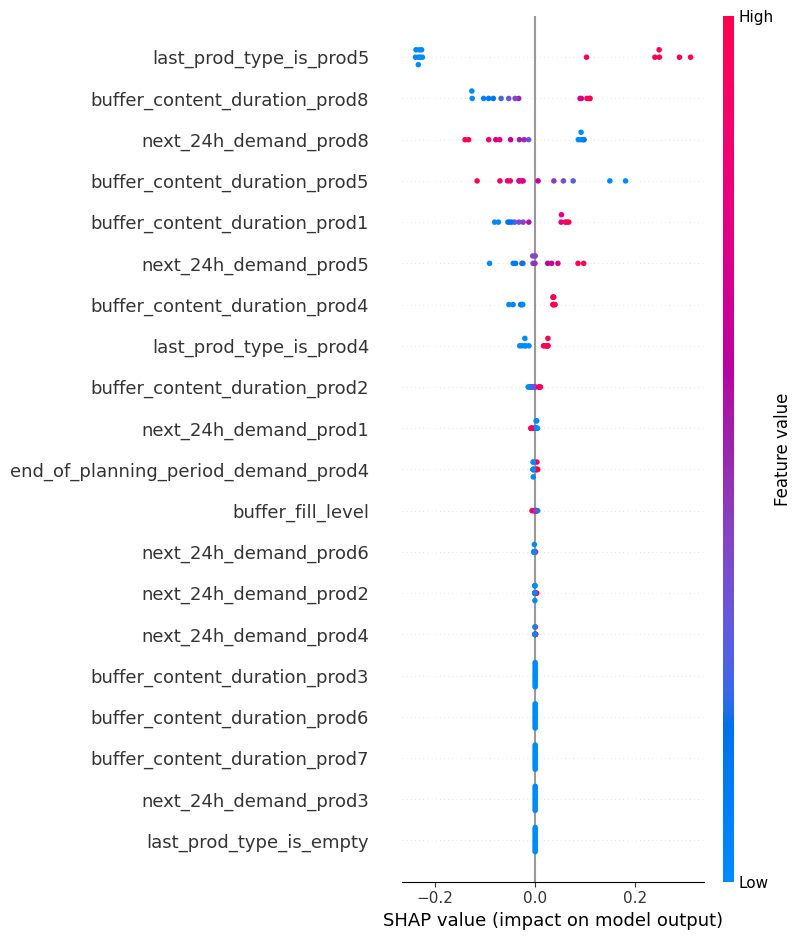}
    \caption{SHAP plot for product 5}
    \label{fig:agg5-week42}
\end{figure}

\begin{table}[ht!]
    \footnotesize
    \centering
    \caption{Top 10 Variables for Input X Gradient prod5}
    \label{tab:top_variables_prod5}
    \begin{tabularx}{\textwidth}{lXr}
        \toprule
        \textbf{Rank} & \textbf{Variable} & \textbf{Value} \\
        \midrule
        1  & end\_of\_planning\_period\_demand\_prod8 & -0.031818896532058716 \\
        2  & last\_prod\_type\_is\_prod4 & -0.026379607617855072 \\
        3  & buffer\_fill\_level & 0.026135722175240517 \\
        4  & buffer\_content\_duration\_prod8 & 0.02558363787829876 \\
        5  & buffer\_content\_duration\_prod7 & 0.020351236686110497 \\
        6  & buffer\_content\_duration\_prod5 & -0.0160413458943367 \\
        7  & buffer\_content\_duration\_prod2 & 0.015320430509746075 \\
        8  & buffer\_content\_duration\_prod1 & -0.015021555125713348 \\
        9  & next\_24h\_demand\_prod8 & -0.012931049801409245 \\
        10 & end\_of\_planning\_period\_demand\_prod5 & 0.011032014153897762 \\
        \bottomrule
    \end{tabularx}
\end{table}
\FloatBarrier

If product 5 was produced last, it is more likely to be produced again (setup-effort-hypothesis).
Fuller buffer content of product 8 makes production of prod5 more likely, while high demand of product 8 speaks against production of product 5 (criticality-hypothesis).
Similarly, less buffer content and more 24h demand of product 5 speak for its production (criticality-hypothesis).

\end{document}